\pdfoutput=1

\documentclass[11pt]{article}

\usepackage[preprint]{nlp_arxiv}

\usepackage{times}
\usepackage{latexsym}

\usepackage[T1]{fontenc}

\usepackage[utf8]{inputenc}

\usepackage{microtype}

\usepackage{inconsolata}

\usepackage{graphicx}

\usepackage{url}            %
\usepackage{booktabs}       %
\usepackage{amsfonts}       %
\usepackage{nicefrac}       %
\usepackage{microtype}      %
\usepackage[table]{xcolor}         %
\usepackage{caption}

\usepackage{mathtools}
\usepackage{multirow}
\usepackage{bm}
\usepackage{epsfig}
\usepackage{graphicx}
\usepackage{subcaption}
\usepackage{amsthm}
\usepackage{amsmath}
\usepackage{amssymb}
\usepackage{enumitem}
\usepackage{makecell}
\usepackage{verbatim}
\usepackage{color}
\usepackage{setspace}
\usepackage{array}
\usepackage{booktabs}
\usepackage{algorithm}
\usepackage[algo2e,algoruled,boxed,lined]{algorithm2e}
\usepackage{algorithmicx}
\usepackage{xcolor}
\usepackage{fix-cm}
\usepackage{arydshln}
\usepackage{colortbl}
\usepackage{wrapfig}
\usepackage{pgfplotstable}
\usepackage{marvosym}

\newcolumntype{a}{>{\columncolor{Gray}}c}

\newcommand{\ie}{\emph{i.e.}}
\newcommand{\eg}{\emph{e.g.}}

\usepackage[pagebackref=true,breaklinks=true,colorlinks=true,bookmarks=false]{hyperref}
\definecolor{deepred}{HTML}{940000}
\hypersetup{linkcolor=deepred}
\hypersetup{urlcolor  = [rgb]{0.4,0.15,0.95}}
\hypersetup{citecolor=[rgb]{0.4,0.15,0.95}}
\usepackage{url}

\usepackage[capitalize,noabbrev]{cleveref}

\usepackage{pifont}
\usepackage{tocloft}
\usepackage[toc,page,header]{appendix}
\usepackage{adjustbox}
\usepackage{minitoc}

\renewcommand \thepart{}
\renewcommand \partname{}

\usepackage[capitalize,noabbrev]{cleveref}

\usepackage{lipsum}

\definecolor{Gray}{gray}{0.94}
\definecolor{darkspringgreen}{rgb}{0.09, 0.45, 0.27}
\definecolor{americanrose}{rgb}{1.0, 0.01, 0.24}

\definecolor{bestcolor}{HTML}{fdbe86}
\definecolor{secondbestcolor}{HTML}{feeddc}

\usepackage{pifont}
\usepackage{tocloft}
\usepackage[toc,page,header]{appendix}
\usepackage{adjustbox}
\usepackage{minitoc}

\renewcommand \thepart{}
\renewcommand \partname{}

\usepackage[linewidth=1pt]{mdframed}

\usepackage{makecell}

\newlength\savewidth\newcommand\shline{\noalign{\global\savewidth\arrayrulewidth
  \global\arrayrulewidth 1pt}\hline\noalign{\global\arrayrulewidth\savewidth}}

\title{Orthogonal Finetuning Made Scalable}

\author{Zeju Qiu\textsuperscript{1,\textdagger}~~~~Weiyang Liu\textsuperscript{1,2,\textdagger,*}~~~~Adrian Weller\textsuperscript{3,4}~~~~Bernhard Sch\"olkopf\textsuperscript{1}\\[2mm]
  \textsuperscript{1}Max Planck Institute for Intelligent Systems~~~~\textsuperscript{2}The Chinese University of Hong Kong\\[0.5mm]
  \textsuperscript{3}University of Cambridge~~~~\textsuperscript{4}The Alan Turing Institute~~~~\textsuperscript{\textdagger}Equal contribution\\[0.5mm]
  \textsuperscript{*}Project lead,~Correspondence to \texttt{wyliu@cse.cuhk.edu.hk}
  \\[3.85mm]
  \href{https://spherelab.ai/oftv2/}{\tt spherelab.ai/oftv2}
  }

\begin{document}
\maketitle

\doparttoc 
\faketableofcontents

\begin{abstract}
Orthogonal finetuning (OFT) offers highly parameter-efficient adaptation while preventing catastrophic forgetting, but its high runtime and memory demands limit practical deployment. We identify the core computational bottleneck in OFT as its weight-centric implementation, which relies on costly matrix-matrix multiplications with cubic complexity. To overcome this, we propose OFTv2, an input-centric reformulation that instead uses matrix-vector multiplications (\ie, matrix-free computation), reducing the computational cost to quadratic. We further introduce the Cayley-Neumann parameterization, an efficient orthogonal parameterization that approximates the matrix inversion in the Cayley transform via a truncated Neumann series. These modifications allow OFTv2 to achieve up to 10$\times$ faster training and 3$\times$ lower GPU memory usage without compromising performance. In addition, we extend OFTv2 to support finetuning quantized foundation models and show that it outperforms the popular QLoRA in training stability, efficiency, and memory usage.
\end{abstract}

\section{Introduction}

As foundation models continue to improve in performance, recent years have witnessed a paradigm shift from end-to-end learning to a pretraining-finetuning framework. This shift underscores the need for finetuning methods that are both effective and scalable. Owing to its training stability and adaptation efficiency, orthogonal finetuning (OFT)~\cite{qiu2023controlling,liu2024boft} has emerged as a promising approach for adapting foundation models to downstream tasks. However, while performing well, OFT incurs high computational and memory costs, limiting its scalability.
Motivated by these challenges, we seek to make OFT more scalable to large foundation models.

\begin{figure}[t]
\centering
    \setlength{\abovecaptionskip}{6pt}
    \setlength{\belowcaptionskip}{-6pt}
    \vspace{-2mm}
\includegraphics[width=1\linewidth]{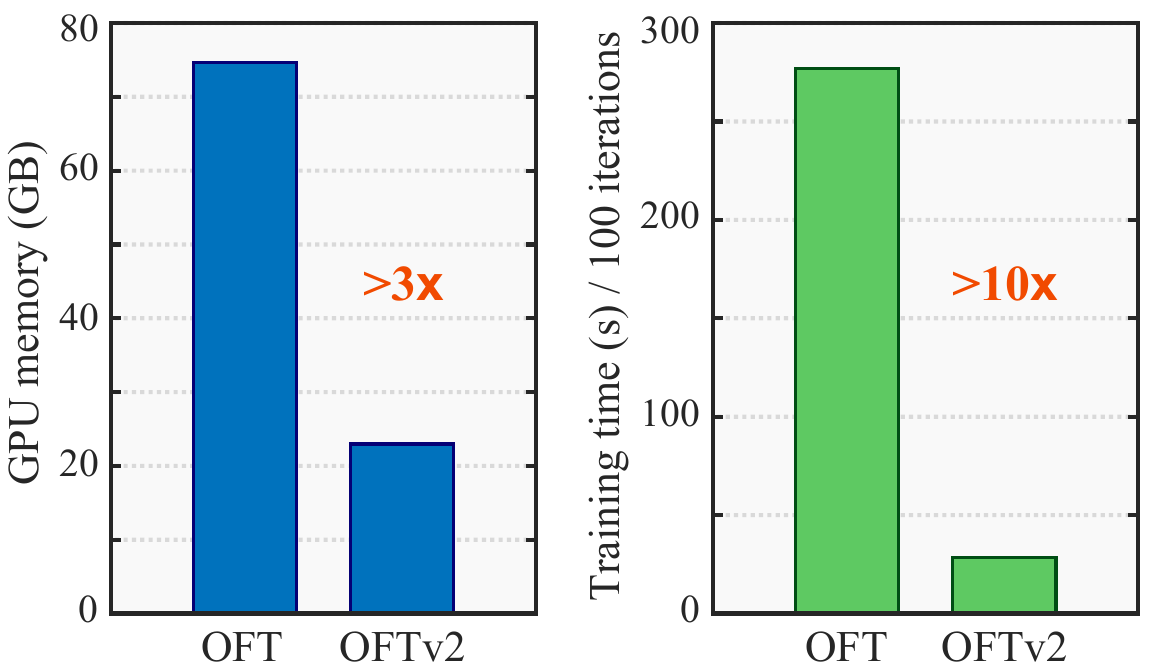}
    \caption{OFTv2 significantly reduces training time and GPU memory usage without sacrificing performance. The finetuning is performed with Qwen2.5-7B.}
    \label{fig:teaser}
\end{figure}

Towards this goal, we begin by identifying the key bottleneck that limits OFT's scalability. At its core, OFT learns layer-shared orthogonal matrices to transform pretrained weight matrices, resulting in a naive \emph{weight-centric} implementation where forward inference is performed after merging the learned orthogonal matrices into weight matrices during training. The weight-centric implementation %
thus involves matrix-matrix multiplications with cubic complexity. As weight matrices grow large, this cubic scaling severely limits OFT's applicability to large foundation models. 
However, these matrix-matrix multiplications are not fundamentally necessary. We draw inspiration from matrix-free methods~\cite{chen2005matrix}, such as the power method and the Lanczos algorithm, which avoid explicit matrix-matrix operations by treating matrices as linear operators applied to vectors. These methods operate entirely through matrix-vector multiplications, applying a matrix to vectors in the appropriate space without ever forming full matrix products. Guided by the same insight, we introduce an \emph{input-centric} implementation of OFT, in which the learned orthogonal transformations are applied directly to the input vectors during each forward pass, rather than being merged into the weight matrix. This reformulation reduces the complexity from cubic to quadratic. We refer to this new formulation as OFTv2. Despite its simplicity, this change significantly enhances the scalability of OFT, making it suitable for finetuning large foundation models that the original OFT could not handle due to memory constraints.

Another scalability bottleneck in OFT arises from the Cayley parameterization used by \citet{liu2021orthogonal,qiu2023controlling,liu2024boft} to preserve orthogonality. While effective, this parameterization involves computing a matrix inverse, which becomes increasingly costly and less numerically stable as weight matrices get larger. To address this, we use a numerically stable yet efficient approximation -- the Cayley–Neumann parameterization (CNP)~\cite{qiu2025poet}. By replacing the matrix inverse in the original Cayley transform with a truncated Neumann series, CNP offers improved numerical stability and lower computational cost, particularly in settings where OFT is applied to finetune large foundation models. With CNP, OFTv2 becomes even more scalable and readily applicable for efficient adaptation of such models. In Figure~\ref{fig:teaser}, we compare OFT and OFTv2 by performing finetuning tasks on Qwen2.5-7B, which is the largest model that the original OFT can finetune within a single Nvidia H100 (80GB). These empirical results demonstrate that OFTv2 achieves substantial GPU memory savings and training speed-up over the original OFT formulation~\cite{qiu2023controlling}.

In practice, finetuning ultra-large foundation models (\eg, LLaMA 3.1-70B~\cite{grattafiori2024llama}, Qwen 2.5-72B~\cite{yang2024qwen2}) typically requires quantization to fit within GPU memory limits. To support this, we follow the general design of the QLoRA framework~\cite{dettmers2023qlora} but replace LoRA with OFTv2. Our input-centric implementation of orthogonal finetuning enables a seamless application to the finetuning of quantized foundation models, resulting in QOFT--an efficient orthogonal finetuning that enables efficient adaptation of quantized ultra-large models.
Our major contributions are summarized below:

\vspace{0.5mm}

\begin{itemize}[leftmargin=4mm,nosep]
\setlength\itemsep{0.4em}

\item Inspired by matrix-free methods that avoid matrix-matrix multiplications in solving linear systems, we propose OFTv2--an input-centric reformulation of OFT that achieves significantly better scalability, with more than 10$\times$ faster training and 3$\times$ lower GPU memory usage.

\item We apply the Cayley–Neumann parameterization~\cite{qiu2025poet} in OFTv2. It approximates the Cayley transform with a truncated Neumann series and eliminates matrix inversions.

\item Owing to the new input-centric formulation, we adapt OFTv2 to finetuning quantized foundation models. This enables memory-efficient finetuning of ultra-large models.

\item We apply OFTv2 and its quantized variant to different foundation models (including large language models and text-to-image generative models) across various model scales.

\end{itemize}

\section{Related Work}

\noindent\textbf{Parameter-efficient finetuning (PEFT)}. As foundation models become increasingly large and powerful, there has been growing interest in finetuning them for downstream tasks in a parameter-efficient manner~\cite{houlsby2019parameter,aghajanyan2020intrinsic,hulora2022,edalati2022krona,wang2022adamix,gheini2021cross,zaken2022bitfit,guo2020parameter,sung2021training,ansell2022composable,lester2021power,li2021prefix,vu2022spot,he2021towards,mao2021unipelt,karimi2021compacter,liu2022few,sung2022lst,chen2022parameter,jia2022visual,chen2022adaptformer,zhang2022neural,jie2023fact,lian2022scaling,luo2023towards,zhang2024autolora,wu2024mixture}. In particular, reparameterization-based methods (\eg, \citet{aghajanyan2020intrinsic,hulora2022,edalati2022krona,zi2023delta,chavan2023one}) are enjoying wide adoption. LoRA~\cite{hulora2022} learns a pair of small low-rank matrices whose product is added to each weight matrix, enabling task adaptation with a small number of trainable parameters. Building on LoRA, several works dynamically adjust the rank across layers to better balance the parameter budget~\cite{zhang2022adaptive,valipour2022dylora,zhang2023increlora,zhang2024autolora}. To improve scalability, QLoRA~\cite{dettmers2023qlora} quantizes the frozen base model to 4-bit NormalFloat with double quantization and back-propagates only through LoRA, achieving near full-precision accuracy while drastically lowering memory usage.

\begin{figure}[t]
\centering
    \setlength{\abovecaptionskip}{4pt}
    \setlength{\belowcaptionskip}{-8pt}
\includegraphics[width=.9\linewidth]{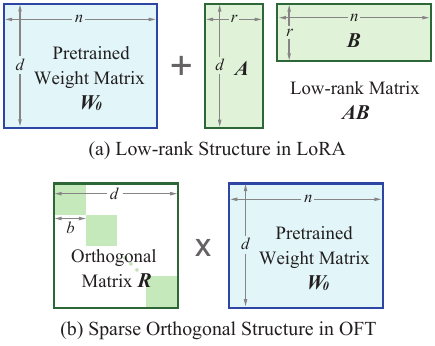}
    \caption{Comparison between LoRA and OFT.}
    \label{fig:lora_oft_comp}
\end{figure}

\noindent\textbf{Orthogonal Finetuning}. \citet{qiu2023controlling,liu2024boft} propose a reparameterization-based method that learns layer-shared orthogonal matrices to transform neurons, yielding strong generalization and stable training. The is motivated by the observation that hyperspherical energy (\ie, a geometric characterization of neurons on the unit sphere) influences generalization~\cite{liu2018learning,liu2021learning,lin2020regularizing,liu2023generalizing}, and that orthogonal transformations keep this energy invariant~\cite{liu2021orthogonal}. A growing body of research~\cite{ma2024parameter,yang2024orthogonal,gorbunov2024group,yuan2024bridging,feng2025omoe,raj2025hyper,lingam2024svft,bini2024ether,su2024defense,liao20243} builds upon the core idea of OFT. Figure~\ref{fig:lora_oft_comp} provides a comparison between OFT and LoRA. OFT achieves parameter efficiency through sparsity, whereas LoRA relies on a low-rank structure.

\section{OFTv2: Faster and More Scalable}

\subsection{Preliminaries}

Let $\bm{W} = [\bm{w}_1, \cdots, \bm{w}_n] \in \mathbb{R}^{d \times n}$ be a weight matrix with columns $\bm{w}_i \in \mathbb{R}^d$. In a linear layer, the forward pass is $\bm{z} = \bm{W}\bm{x}$, where $\bm{x} \in \mathbb{R}^d$ is the input and $\bm{z} \in \mathbb{R}^n$ is the output. OFT reparameterizes the weight matrix with $\bm{W}_{\text{OFT}}=\bm{R}\bm{W}_0$ where $\bm{W}_0$ is the pretrained weight matrix and $\bm{R}\in\mathbf{R}^{d\times d}$ is an orthogonal matrix. OFT only learns $\bm{R}$ for adapting the pretrained model to downstream tasks. To enforce orthogonality, \citet{liu2021learning,qiu2023controlling,liu2024boft} parameterize $\bm{R}$ using the Cayley transform: $\bm{R} = (\bm{I} + \bm{Q})(\bm{I} - \bm{Q})^{-1}$, where $\bm{Q}$ is a skew-symmetric matrix satisfying $\bm{Q} = -\bm{Q}^\top$. To further improve parameter-efficiency, OFT constrains the orthogonal matrix $\bm{R}$ to have a block-diagonal structure: $\bm{R}=\text{Diag}(\bm{R}_1,\cdots,\bm{R}_r)$ where for any $i$, $\bm{R}_i\in\mathbb{R}^{b\times b}$ is a small orthogonal matrix and $b\cdot r=d$. Each  $\bm{R}_i$ can be parameterized using the Cayley transform. This block-diagonal form imposes a sparsity pattern on $\bm{R}$, effectively making it a sparse orthogonal matrix. Leveraging this structure, \citet{liu2024boft} further enhances parameter efficiency using butterfly factorization.

\subsection{From Weight-centric Implementation to Input-centric Implementation}

OFT performs finetuning by learning an orthogonal matrix to directly transform the weight matrix, which naturally leads to a weight-centric implementation of the forward pass:
\begin{equation}
    \bm{z} = \underbrace{\overbrace{\bm{W}_0^\top\bm{R}^\top}^{\text{(1) \textbf{Weight transform}: matrix-matrix mult.}}\bm{x}}_{\text{(2) \textbf{Linear map}: matrix-vector mult.}}
\end{equation}
The original OFT first performs a weight transform by computing $\bm{W}_{\text{OFT}}^\top=\bm{W}_0^\top\bm{R}^\top$ 
(\ie, a matrix-matrix multiplication) and then computes the results of a linear layer with the equivalent weight matrix $\bm{W}_{\text{OFT}}^\top$ (\ie, a matrix-vector multiplication). This incurs $\mathcal{O}(nd^2)$ complexity due to the matrix-matrix multiplication. Inspired by matrix-free methods for solving linear systems, we observe that OFT’s forward pass can be interpreted as two linear maps applied to the input. This leads to an input-centric implementation
\begin{equation}
    \bm{z} = \underbrace{\bm{W}_0^\top\overbrace{\bm{R}^\top\bm{x}}^{\text{(1) \textbf{Linear map}: matrix-vector mult.}}}_{\text{(2) \textbf{Linear map}: matrix-vector mult.}}
\end{equation}
where only two matrix-vector multiplications are required, reducing the complexity from cubic to quadratic: $\mathcal{O}(nd + d^2)$. This simple conceptual shift in implementation entails a substantial speed-up in training time and reduction in GPU memory.

\subsection{Approximate Orthogonality via Cayley-Neumann Parameterization}

The Cayley parameterization constructs an orthogonal matrix $\bm{R}$ with $(\bm{I} + \bm{Q})(\bm{I} - \bm{Q})^{-1}$, where $\bm{Q}$ is a skew-symmetric matrix. One limitation of this formulation is that it only generates rotation matrices, though empirical studies~\cite{liu2021orthogonal,qiu2023controlling,liu2024boft} suggest that this restriction does not negatively affect performance. More critically, computing a matrix inverse introduces numerical instability and additional computational overhead, making it challenging to scale to large orthogonal matrices. To address this, we use the Cayley-Neumann parameterization proposed by \citet{qiu2025poet}, where the matrix inverse is approximated by a truncated Neumann series:

\vspace{-4.75mm}
\begin{equation*}\label{eq:cnp}
\begin{aligned}
    \bm{R}&=(\bm{I}+\bm{Q})(\bm{I}-\bm{Q})^{-1}=(\bm{I}+\bm{Q})\big(\sum_{i=0}^\infty \bm{Q}^i \big) \\[-2.75mm]
    &\approx (\bm{I}+\bm{Q})\big(\bm{I}+\sum_{i=1}^k \bm{Q}^i \big),
\end{aligned}
\end{equation*}
\noindent where larger $k$ leads to better approximation. Removing the matrix inversion improves training stability. The Neumann series approximation converges in the operator norm if $\|\bm{Q}\|<1$. 
This condition is naturally satisfied in practice: 
to start from the pretrained model, OFT initializes the orthogonal matrix $\bm{R}$ as the identity, which requires $\bm{Q}$ to start as a zero matrix. 
Since finetuning begins with a small learning rate and typically involves relatively few steps, $\bm{Q}$ tends not to drift far from zero. Empirically, even if $\|\bm{Q}\|$ slightly exceeds $1$, it does not harm OFT's training stability, as we use only a finite number of Neumann terms.

\noindent\textbf{Custom CUDA kernel for skew-symmetric matrices}. To maximize GPU memory efficiency, we leverage the skew-symmetric structure of $\bm{Q}\in\mathbb{R}^{n\times n}$, where $Q_{ii} = 0$, $Q_{ij} = -Q_{ji}$. By storing only the upper triangular part as a vector, we reduce the storage requirement from $n^2$ to $\frac{n (n - 1)}{2}$. During the forward pass, $\bm{Q}$ is reconstructed on-the-fly using a highly optimized custom
CUDA kernel that significantly accelerates this process.

\section{QOFT: Adapting OFTv2 to Finetuning Quantized Foundation Models}

While PEFT methods primarily aim to reduce optimizer memory by minimizing trainable parameters, the growing scale of foundation models has shifted the memory bottleneck to the pretrained weights themselves. As model dimensions grow, these frozen parameters increasingly dominate memory consumption during training~\cite{kim2023memoryefficientfinetuningcompressedlarge}. To address this emerging challenge, we argue that truly scalable OFT must operate directly on quantized model representations, such as NormalFloat4~\cite{dettmers2023qlora} and AWQ~\cite{lin2024awqactivationawareweightquantization}. This represents a critical shift that enables OFT to scale effectively.

To this end, we introduce QOFT, a natural extension of OFTv2 for quantized foundation models. QOFT largely follows the framework of QLoRA~\cite{dettmers2023qlora}. Specifically, the quantized low-bit weight matrices are first dequantized to higher precision, after which the parameter-efficient adaptation is carried out in the higher-precision space. Formally, the forward pass of QOFT can be written as
\begin{equation}
    \bm{z} = \underbrace{\text{Dequant}(\bm{W}_{\text{quant}})^\top}_{\text{Fronzen}}\underbrace{\bm{R}^\top}_{\text{Trainable}}\bm{x}
\end{equation}
The update of OFTv2’s orthogonal matrix $\bm{R}$ is performed in high precision (\eg, BF16). We denote the dequantization function as $\text{Dequant}(\cdot)$ and follow QLoRA’s design by adopting a double quantization strategy, where the quantization parameters of the weight matrices are themselves quantized to further reduce GPU memory usage.

\noindent\textbf{Flexible quantized finetuning via OFTv2}. We now explain why the weight-centric implementation of OFT is ill-suited for quantized foundation models. Computing the matrix product $\bm{W}_{\text{quant}}^\top \bm{R}^\top$ involves rotating (or reflecting) a quantized weight matrix, which requires first dequantizing it to higher precision before applying the transformation. While this is mathematically valid, it makes OFT dependent on the specific quantization method used. Different quantization schemes may require different treatments for computing $\text{Dequant}(\bm{W}_{\text{quant}})^\top \bm{R}^\top$, introducing unnecessary complexity. In contrast, the input-centric implementation avoids this issue by fully decoupling OFT from weight quantization. It applies the learned orthogonal matrix $\bm{R}^\top$ to the input $\bm{x}$.
The subsequent forward pass proceeds as usual under any quantization strategy. As a result, OFTv2 becomes a quantization-agnostic PEFT method compatible with arbitrary weight quantization schemes.

\noindent\textbf{QOFT vs. QLoRA}. We now look into the forward pass of QLoRA: $\bm{z}=\text{Dequant}(\bm{W}_{\text{quant}})^\top\bm{x}+(\bm{A}\bm{B})^\top \bm{x}$ where $\bm{A}\in\mathbb{R}^{d\times r}$ and $\bm{B}\in\mathbb{R}^{r\times n}$ are low-rank matrices and $r \ll \min(d,n)$ is usually quite small. First, QOFT is more suitable for post-training quantization when merging the finetuned weights back into the quantized model. In QLoRA, the equivalent weight $\bm{W} + \bm{A}\bm{B}$ can alter the dynamic range (\ie, the possible minimum and maximum values) of the weight matrix, potentially complicating requantization. In contrast, the equivalent weight in QOFT, $\bm{R}\bm{W}$, preserve the dynamic range of individual elements. The worse-case requantization error for QLoRA is always larger than QOFT by $\|\bm{A}\bm{B}\|_{\infty}$. This advantage is also partially supported by recent evidence~\cite{tseng2024quip,ashkboos2024quarot} suggesting that orthogonal transformations can homogenize weight magnitudes and suppress outliers.

Another practical limitation of QLoRA is its training instability. Across various experiments, we observe that QLoRA is prone to loss divergence and unstable optimization. We suspect this arises from the inherently noisier gradients in QLoRA, which adversely affect the finetuned weights. In contrast, QOFT benefits from the orthogonality of $\bm{R}$, which also regularizes the back-propagated gradients. As a result, the adaptation weights in QOFT are better conditioned, and when merged into the pretrained model, they yield a more stable finetuned model. This observation is supported by prior work~\cite{qiu2023controlling,liu2024boft} showing that OFT significantly improves training stability and mitigates catastrophic forgetting.

\begin{figure}[t]
\centering
    \setlength{\abovecaptionskip}{6pt}
    \setlength{\belowcaptionskip}{-8pt}
\includegraphics[width=.99\linewidth]{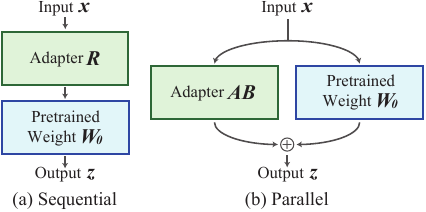}
        \caption{Comparison between sequential (\eg, OFT) and parallel (\eg, LoRA) adaptation.}
    \label{fig:seq_para_comp}
\end{figure}

\section{Discussions and Intriguing Insights}

\noindent\textbf{Sparse vs.\ low-rank PEFT}. As shown in Figure~\ref{fig:lora_oft_comp}, OFT and LoRA achieve parameter-efficiency through sparsity and low rank, respectively. This suggests an intriguing analogy between OFT and LoRA, as sparsity and low rank represent arguably two of the most widely studied and exploited structural properties in matrices. To further enhance the scalability of OFT, more structured sparsity should be exploited, \eg, butterfly factorization~\cite{liu2024boft}. Moreover, similar to AdaLoRA~\cite{zhang2023adalora}, the sparsity level in OFT can be conditioned on the task and layer. Compared to low-rank PEFT, sparse PEFT approaches like OFT remain relatively underexplored, leaving many interesting open problems for future investigation.

\noindent\textbf{Sequential vs.\ parallel adaptation}. As shown in Figure~\ref{fig:seq_para_comp}, OFT and LoRA exemplify two distinct adaptation strategies: sequential adaptation and parallel adaptation, respectively. This contrast is particularly intriguing, as it explains why sequential adaptation benefits from orthogonality, while parallel adaptation naturally aligns with low rank. Sequential adaptation offers great expressiveness but is also more susceptible to error propagation and distortion of the pretrained model’s spectral properties. Enforcing orthogonality on $\bm{R}$ is therefore a natural choice, as it preserves these properties and helps prevent the accumulation of errors. Sparsity is the natural choice if we want to save parameters in orthogonal matrices. Parallel adaptation adds the adapter $\bm{R}$ to the pretrained model.
In this case, we want $\bm{R}$ to be a dense update while maintaining parameter efficiency--a goal naturally achieved through low-rank matrices. This perspective may inspire new directions in adapter design.

\noindent\textbf{Efficient orthogonality parameterization}. OFT also highlights the importance of efficient parameterization of orthogonal matrices. In fact, the efficiency is closely tied to two factors: (1) the degree to which orthogonality needs to be approximated, and (2) the size of the set of orthogonal matrices considered. Our experiments indicate that exact orthogonality and the full orthogonal group are not strictly necessary, as parameterizations from the special orthogonal group and approximate orthogonality perform quite well in practice. This raises an open question: can we find even more efficient parameterizations with comparable performance?

\begin{figure*}[t]
\centering
    \setlength{\abovecaptionskip}{6pt}
    \setlength{\belowcaptionskip}{-3pt}
\includegraphics[width=\linewidth]{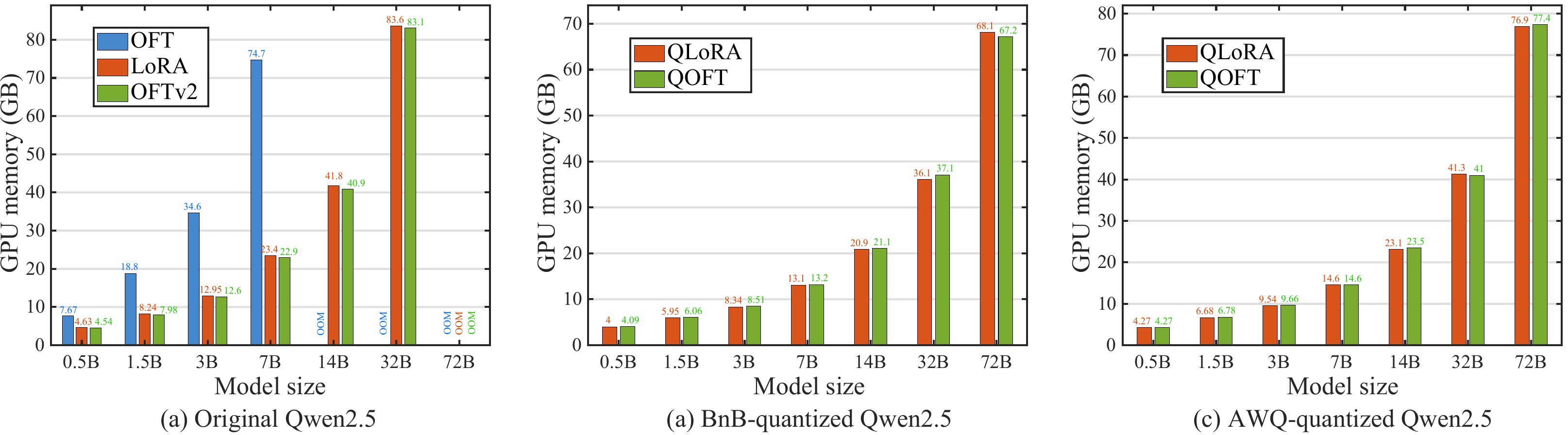}
        \caption{Results of GPU memory usage for the same finetuning task. (a) OFT, LoRA and OFTv2 on Qwen2.5; (b) QLoRA and QOFT on NF4-quantized Qwen2.5; (c) QLoRA and QOFT on AWQ-quantized Qwen2.5.}
    \label{fig:mem_comp}
\end{figure*}

\section{Experiments on Scalability}

Our experiments systematically evaluate OFTv2 along two key dimensions: (1) its scalability improvements over the original OFT, and (2) its finetuning performance across a diverse set of tasks from multiple domains. For both aspects, we compare OFTv2 and QOFT against the well-established, memory- and compute-efficient low-rank adaptation methods LoRA~\cite{hu2022lora} and QLoRA~\cite{dettmers2023qlora}.

\subsection{GPU Memory Efficiency}

As depicted in Figure~\ref{fig:teaser}, OFTv2 achieves a $3\times$ reduction in \textbf{GPU memory consumption} compared to the original OFT when finetuning the Qwen2.5-7B model. Furthermore, QOFT significantly reduces memory consumption by enabling the orthogonal finetuning of quantized base models. In the following ablation studies comparing against both LoRA and QLoRA baselines, where QLoRA broadly refers to low-rank adaptation of quantized models without being limited to NormalFloat 4-bit quantization, we evaluate the actual GPU memory consumption during finetuning of Qwen2.5 models from 0.5B to 72B parameters. For a comprehensive analysis, we additionally incorporate the widely adopted quantization method AWQ~\cite{lin2024awqactivationawareweightquantization} for activation-aware quantization. The results are summarized in Figure~\ref{fig:mem_comp}. Our experimental results demonstrate that OFTv2 and QOFT achieve memory efficiency comparable to low-rank adaptation methods, with a consistent performance across model scales and data formats.

\subsection{Computational Efficiency}

We begin by evaluating the training speed of OFTv2 relative to the original OFT. To this end, we finetune a Qwen2.5-7B model on the OASST1-Guanaco-9K dataset~\cite{dettmers2023qlora} for instruction following and measure the training time. As shown in Figure~\ref{fig:teaser}, OFTv2 achieves a 3$\times$ speed-up over the original OFT. 
We further compare the overall training speed of OFTv2 and LoRA across different model scales and precisions.
Settings from both the GSM8K experiment (Table~\ref{tab:llama}) and the OpenR1-Math-220k experiment~\cite{open-r1} (Table~\ref{tab:math_main}) are used for comparison. Clock times for each setting are reported in Table~\ref{tab:training_time} and Table~\ref{tab:training_time_q}.
While low-rank adaptation methods like LoRA benefit from PyTorch's highly optimized GEMM operations via NVIDIA cuBLAS/cuDNN libraries, the simple designs in OFTv2 significantly narrow this optimization gap in full-precision settings. Notably, OFTv2 outperforms LoRA in quantized settings (Table~\ref{tab:training_time_q}), demonstrating that its quantization-agnostic design effectively leverages underlying quantization-layer optimizations.

\begin{table}[t!]
\small
\centering
\setlength{\tabcolsep}{8.5pt}
\renewcommand{\arraystretch}{1.35}
\setlength{\abovecaptionskip}{6pt}
\setlength{\belowcaptionskip}{-3pt}
\begin{tabular}{lccc}
\textbf{Model Size} & \textbf{GPUs} & \textbf{LoRA} &  \cellcolor{Gray}\textbf{OFTv2} \\\shline
Llama-2-7B   & 8$\times$H100 & \textbf{00:12:10} &\cellcolor{Gray} 00:15:10 \\
Llama-2-13B  & 8$\times$H100 & \textbf{00:17:00} &\cellcolor{Gray} 00:19:50 \\
\end{tabular}
\caption{Training time (clock time) comparison: OFTv2 vs. LoRA on GSM8K for mathematical reasoning.}
\label{tab:training_time}
\end{table}

\section{Experiments on Performance}

Having established that OFTv2 achieves comparable memory and computational efficiency to low-rank adaptation methods, we then test its performance on a variety of tasks.

\subsection{Encoder-Decoder Model: BART} 
We evaluate the finetuning of BART-large~\cite{lewis2019bartdenoisingsequencetosequencepretraining} on the XSum~\cite{narayan2018dontdetailsjustsummary} and CNN/DailyMail~\cite{hermann2015teachingmachinesreadcomprehend} datasets for text summarization, reporting ROUGE-1/2/L scores for LoRA and OFTv2 under both full-precision and NormalFloat4 4-bit quantization.
We further investigate different configurations by increasing the rank $r$ for LoRA and the block size $b$ for OFTv2. The results from these finetuning tasks are reported in Table~\ref{tab:bart_summerization}. We observe that OFTv2/QOFT consistently outperforms LoRA/QLoRA across all tested configurations, while notably utilizing 47–53\% fewer trainable parameters. The performance gain gets more obvious with increasing model capacity: at the maximum parameter budget, QOFT outperforms QLoRA by +0.93 ROUGE-1 on XSum (44.16 vs. 43.23), suggesting a more effective utilization of expanded adapters. Furthermore, the finetuning performance of OFTv2/QOFT further improves with an increase budget of trainable parameters.

\begin{table}[t!]
\small
\centering
\setlength{\tabcolsep}{7.5pt}
\renewcommand{\arraystretch}{1.35}
\setlength{\abovecaptionskip}{6pt}
\setlength{\belowcaptionskip}{-6pt}
\begin{tabular}{lccc}
\textbf{Model Size} & \textbf{GPUs} & \textbf{QLoRA} & \cellcolor{Gray} \textbf{QOFT} \\
\shline
Qwen2.5-1.5B & 8$\times$H100 & 01:20:00 & \cellcolor{Gray} \textbf{01:17:30} \\
Qwen2.5-7B   & 8$\times$H100 & 03:25:00 & \cellcolor{Gray} \textbf{03:19:30} \\
Qwen2.5-32B  & 8$\times$H100 & 12:51:45 & \cellcolor{Gray} \textbf{12:27:45} \\
\end{tabular}
\caption{Clock time comparison of QOFT and QLoRA on OpenR1-Math-220k for mathematical reasoning.}
\label{tab:training_time_q}
\end{table}

\begin{figure*}[t!]
\centering
    \setlength{\abovecaptionskip}{4pt}
    \setlength{\belowcaptionskip}{6pt}
\includegraphics[width=1\linewidth]{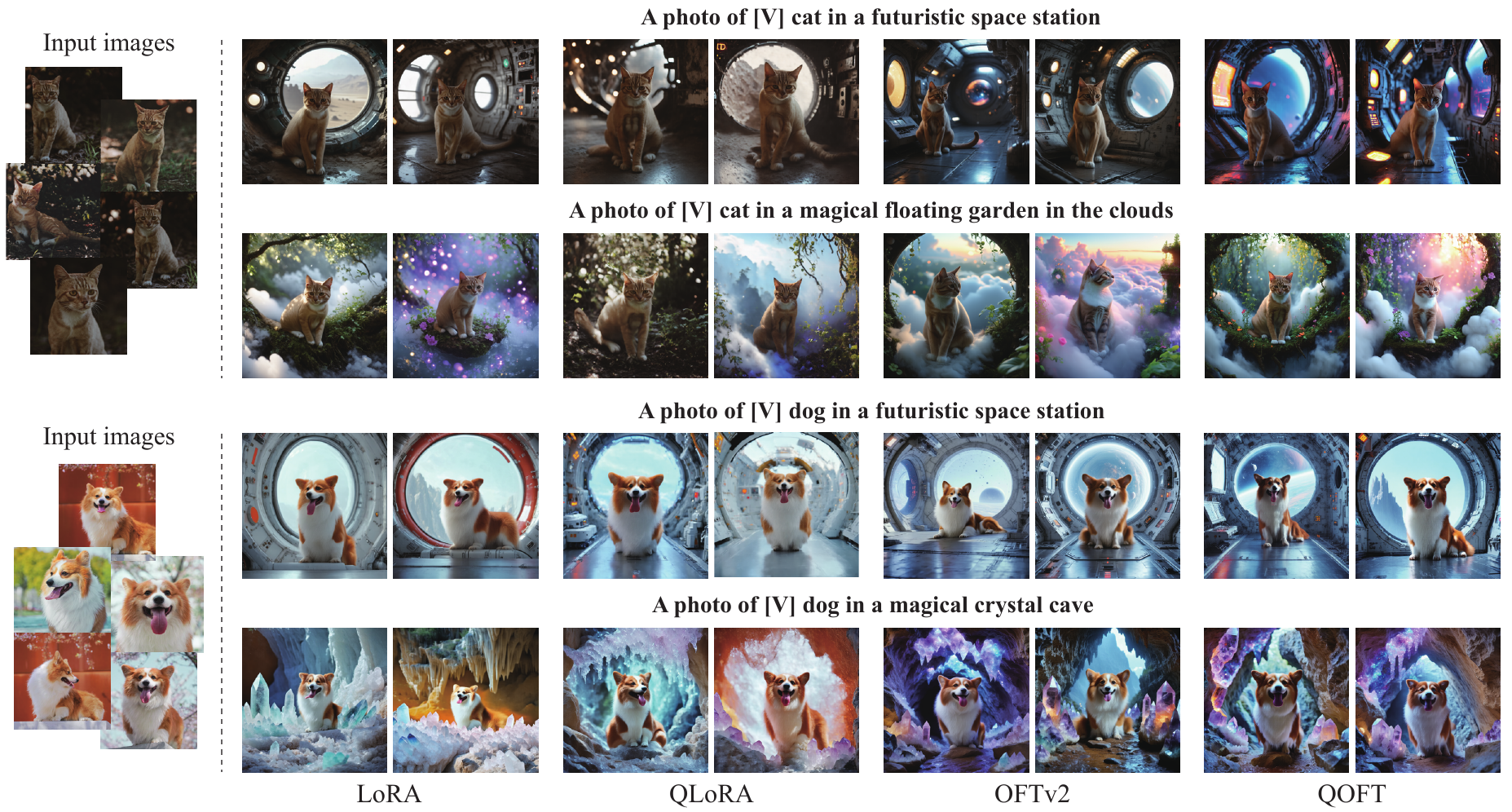}
        \caption{
    Qualitative results from Dreambooth finetuning of Stable Diffusion 3.5 Large (8.1B parameters), with peak allocated GPU memory: LoRA (\textbf{52.33 GB}), OFT (\textbf{52.32 GB}), QLoRA (\textbf{41.60 GB}) and QOFT (\textbf{41.53 GB}).}\label{fig:sd3_large}
\end{figure*}

\begin{table*}[t!]
\centering
\small
\setlength{\tabcolsep}{3.1pt}
\renewcommand{\arraystretch}{1.35}
\setlength{\abovecaptionskip}{6pt}
    \setlength{\belowcaptionskip}{-3pt}
\begin{tabular}{c|ccc|ccc}
\multirow{2}{*}{\bf Quant.} & \multicolumn{3}{c|}{\bf LoRA / QLoRA} &  \multicolumn{3}{c}{\bf OFTv2 / QOFT} \\
& {\bf \# Params} & {\bf XSum$\uparrow$} & {\bf CNN/DailyMail$\uparrow$} &{\bf \# Params} & {\bf XSum$\uparrow$} & {\bf CNN/DailyMail$\uparrow$} \\
\shline
\multirow{3}{*}{Full Prec.} & 4.33M & 43.33 / 20.06 / 35.11 & 43.11 / 20.22 / 29.69 &\cellcolor{Gray} 2.03M &\cellcolor{Gray} \textbf{43.36} / \textbf{20.21} / \textbf{35.31} & \cellcolor{Gray}\textbf{43.27} / \textbf{20.29} / \textbf{29.71} \\
& 8.65M & 43.47 / 20.19 / 35.21 & 43.20 / 20.31 / 29.71 &\cellcolor{Gray} 4.19M & \cellcolor{Gray} \textbf{43.85} / \textbf{20.69} / \textbf{35.83} &\cellcolor{Gray} \textbf{43.72} / \textbf{20.73} / \textbf{30.22} \\
& 17.30M & 43.38 / 20.20 / 35.25 & 43.17 / 20.31 / 29.72 &\cellcolor{Gray} 8.52M & \cellcolor{Gray} \textbf{44.12} / \textbf{20.96} / \textbf{36.01} & \cellcolor{Gray} \textbf{44.08} / \textbf{21.02} / \textbf{30.68} \\
\hline
\multirow{3}{*}{NF4} & 4.33M & 43.09 / 19.82 / 34.92 & 43.17 / 20.25 / 29.66 &\cellcolor{Gray} 2.03M &\cellcolor{Gray} \textbf{43.10} / \textbf{19.92} / \textbf{35.00} &\cellcolor{Gray} \textbf{43.31} / \textbf{20.37} / \textbf{29.74} \\
& \multirow{1}{*}{8.65M} & 43.15 / 19.80 / 34.92 & 43.10 / 20.24 / 29.65 &\cellcolor{Gray} 4.19M &\cellcolor{Gray} \textbf{43.72} / \textbf{20.58} / \textbf{35.68} &\cellcolor{Gray} \textbf{43.71} / \textbf{20.74} / \textbf{30.22} \\
& \multirow{1}{*}{17.30M} & 43.23 / 19.92 / 35.10 & 43.11 / 20.23 / 29.63 &\cellcolor{Gray} 8.52M &\cellcolor{Gray} \textbf{44.16} / \textbf{20.98} / \textbf{36.09} &\cellcolor{Gray} \textbf{44.10} / \textbf{21.05} / \textbf{30.69}
\end{tabular}
\caption{ROUGE-1, ROUGE-2, and ROUGE-L scores for BART-large finetuned on XSum and CNN/DailyMail.}
\label{tab:bart_summerization}
\end{table*}

\begin{table}[t!]
\scriptsize
\setlength{\tabcolsep}{5.75pt}
\renewcommand{\arraystretch}{1.35}
\centering
\setlength{\abovecaptionskip}{6pt}
\setlength{\belowcaptionskip}{-8pt}
\begin{tabular}{c|lcccc}
\multirow{2}{*}{\bf Model} & \multirow{2}{*}{\bf Metric} & \multicolumn{2}{c}{\bf 16-bit} & \multicolumn{2}{c}{\bf 4-bit} \\
& & {\bf LoRA} & \cellcolor{Gray} {\bf OFTv2} & {\bf QLoRA} & \cellcolor{Gray} {\bf QOFT} \\\shline
\multirow{3}{*}{\textbf{7B}} & {\bf\# Params} & 39.98M & \cellcolor{Gray} 17.65M & 39.98M & \cellcolor{Gray} 17.65M \\
& WikiText-2$\downarrow$ & 6.63 & \cellcolor{Gray} \textbf{6.14} & 5.74 & \cellcolor{Gray} \textbf{5.60} \\
& GSM8K$\uparrow$ & 33.81 & \cellcolor{Gray} \textbf{34.65} & 34.12 & \cellcolor{Gray} \textbf{37.23} \\\hline
\multirow{3}{*}{\textbf{13B}} & {\bf\# Params} & 62.59M & \cellcolor{Gray} 27.62M & 62.59M & \cellcolor{Gray} 27.62M \\
& WikiText-2$\downarrow$ & 5.23 & \cellcolor{Gray} \textbf{4.98} & 5.31 & \cellcolor{Gray} \textbf{5.05} \\
& GSM8K$\uparrow$ & 45.94 & \cellcolor{Gray} \textbf{46.02} & 44.20 & \cellcolor{Gray} \textbf{47.92} \\
\end{tabular}
\caption{Finetuning results of Llama-2 models on WikiText-2 (perplexity) and GSM8K (test accuracy).}
\label{tab:llama}
\end{table}

\begin{table*}[t!]
\small
\centering
\setlength{\tabcolsep}{5.25pt}
\renewcommand{\arraystretch}{1.35}
\setlength{\abovecaptionskip}{6pt}
\setlength{\belowcaptionskip}{-5pt}
\begin{tabular}{l|lcccccccc}
\multirow{2}{*}{\textbf{Model}} & \multirow{2}{*}{\textbf{Type}} & \multirow{2}{*}{\textbf{\# Params}} & \multirow{2}{*}{\textbf{AMC23}} & \multirow{2}{*}{\textbf{AQUA}} & \multirow{2}{*}{\textbf{CMATH}} & \textbf{GaoKao} & \textbf{Minerva} & \textbf{Olympiad/} & \textbf{SAT} \\
 & & & & & & \textbf{2023 En} & \textbf{Math} & \textbf{Bench} & \textbf{Math} \\
\shline
\multirow{3}{*}{\textbf{Qwen2.5-1.5B-it}} & \textbf{Baseline} & - & 17.5 & 49.2 & 65.2 & 36.4 & 9.6 & 12.0 & 59.4 \\
 & \textbf{QLoRA} & 18.46M & 15.0 & 42.5 & 61.5 & 29.6 & 8.1 & 8.9 & 59.4 \\
 & \cellcolor{Gray} \textbf{QOFT} &\cellcolor{Gray} 7.89M &\cellcolor{Gray} \textbf{27.5} &\cellcolor{Gray} \textbf{53.1} &\cellcolor{Gray} \textbf{68.5} &\cellcolor{Gray} \textbf{41.0} &\cellcolor{Gray} \textbf{11.8} &\cellcolor{Gray} \textbf{14.4} &\cellcolor{Gray} \textbf{81.2} \\
\hline
\multirow{3}{*}{\textbf{Qwen2.5-1.5B}} & \textbf{Baseline} & -  & 0.0 & 18.9 & 4.0 & 4.2 & 2.6 & 2.4 & 28.1 \\
& \textbf{QLoRA} & 18.46M & 15.0 & 37.4 & \textbf{64.2} & 26.8 & \textbf{8.5} & 6.8 & 62.5 \\
&\cellcolor{Gray} \textbf{QOFT} &\cellcolor{Gray} 7.89M &\cellcolor{Gray} \textbf{22.5} &\cellcolor{Gray} \textbf{53.1} &\cellcolor{Gray} 56.3 &\cellcolor{Gray} \textbf{36.1} &\cellcolor{Gray} \textbf{8.5} &\cellcolor{Gray} \textbf{12.7} &\cellcolor{Gray} \textbf{87.5} \\
\hline
\multirow{3}{*}{\textbf{Qwen2.5-7B-it}} & \textbf{Baseline} & - & 50.0 & 16.5 & 89.3 & 61.8 & \textbf{33.5} & 36.6 & 53.1 \\
& \textbf{QLoRA} & 40.37M & 30.0 & 48.0 & 88.8 & 50.1 & 25.4 & 19.7 & 68.8 \\
&\cellcolor{Gray} \textbf{QOFT} &\cellcolor{Gray} 17.55M &\cellcolor{Gray} \textbf{52.5} &\cellcolor{Gray} \textbf{70.9} &\cellcolor{Gray} \textbf{90.5} &\cellcolor{Gray} \textbf{63.6} &\cellcolor{Gray} \textbf{33.5} &\cellcolor{Gray} \textbf{37.6} &\cellcolor{Gray} \textbf{96.9} \\
\hline
\multirow{3}{*}{\textbf{Qwen2.5-7B}} & \textbf{Baseline} & - & 25.0 & 55.1 & 61.2 & 42.9 & 11.8 & 29.9 & 71.9 \\
& \textbf{QLoRA} & 40.37M & 35.0 & 48.8 & 73.7 & 49.9 & 18.8 & 18.5 & 62.5 \\
&\cellcolor{Gray} \textbf{QOFT} &\cellcolor{Gray} 17.55M &\cellcolor{Gray} \textbf{52.5} &\cellcolor{Gray} \textbf{59.4} &\cellcolor{Gray} \textbf{80.7} &\cellcolor{Gray} \textbf{55.6} &\cellcolor{Gray} \textbf{21.7} &\cellcolor{Gray} \textbf{34.7} &\cellcolor{Gray} \textbf{87.5} \\
\hline
\multirow{3}{*}{\textbf{Qwen2.5-32B-it}} & \textbf{Baseline} & - & 62.5 & 18.5 & 92.5 & 70.1 & \textbf{41.5} & 44.4 & 65.6 \\
& \textbf{QLoRA} & 134.22M & 62.5 & 71.7 & 94.0 & 71.2 & 39.7 & 46.8 & 96.9 \\
&\cellcolor{Gray} \textbf{QOFT} &\cellcolor{Gray} 57.90M &\cellcolor{Gray} \textbf{75.0} &\cellcolor{Gray} \textbf{83.1} &\cellcolor{Gray} \textbf{94.7} &\cellcolor{Gray} \textbf{73.5} &\cellcolor{Gray} \textbf{41.5} &\cellcolor{Gray} \textbf{48.7} &\cellcolor{Gray} \textbf{100.0} \\
\hline
\multirow{3}{*}{\textbf{Qwen2.5-32B}} & \textbf{Baseline} & - & 35.0 & 23.2 & 35.7 & 46.8 & 20.2 & 25.2 & 62.5 \\
& \textbf{QLoRA} & 134.22M & 40.0 & 52.4 & 90.5 & 61.0 & 32.0 & 29.8 & 65.6 \\
&\cellcolor{Gray} \textbf{QOFT} &\cellcolor{Gray} 57.90M &\cellcolor{Gray} \textbf{70.0} &\cellcolor{Gray} \textbf{68.5} &\cellcolor{Gray} \textbf{90.7} &\cellcolor{Gray} \textbf{71.4} &\cellcolor{Gray} \textbf{36.0} &\cellcolor{Gray} \textbf{44.9} &\cellcolor{Gray} \textbf{93.8} \\
\end{tabular}
\caption{Pass@1 performance of the Qwen2.5 series LLMs and its QLoRA/QOFT finetuned variants using the chain-of-thought reasoning distilled from DeepSeek R1.}
\label{tab:math_main}
\end{table*}

\subsection{Decoder-only Model: Llama-2 Series}
We finetune Llama-2 7B and 13B models on the NLG datasets GSM8K~\cite{cobbe2021training} and WikiText-2~\cite{merity2016pointersentinelmixturemodels}. To ensure fairness, we use the same set of hyperparameters for each method across datasets, precisions, and model scales. Both LoRA and QLoRA set rank to $r=16$. Both OFTv2 and QOFT set block size to $b=32$. Table~\ref{tab:llama} shows that OFTv2 consistently outperforms the low-rank adapter across different settings.

\subsection{Decoder-only Model: Qwen2.5 Series}

We perform supervised finetuning on the Huggingface OpenR1-Math-220k~\cite{open-r1} dataset—a large-scale mathematical reasoning corpus containing challenging problems and two to four reasoning traces distilled from DeepSeek R1~\cite{deepseekai2025deepseekr1incentivizingreasoningcapability}. Following the evaluation protocol of Qwen2.5-Math~\cite{yang2024qwen2}, we report pass@1 performance on established math benchmarks: CMATH~\cite{wei2023cmathlanguagemodelpass}, AMC23~\cite{amc23}, AQUA~\cite{ling2017programinductionrationalegeneration}, Olympiad Bench~\cite{he2024olympiadbenchchallengingbenchmarkpromoting}, Gaokao 2023 En~\cite{liao2024mariomathreasoningcode}, and Minerva Math~\cite{lewkowycz2022solvingquantitativereasoningproblems}.
Finetuning was only performed on NormalFloat 4-bit quantized base models due to the substantial memory requirements imposed by the large context window size (16384), necessary for training on a reasoning dataset. The results are reported in Table~\ref{tab:math_main}. The baseline method refers to the pre-trained Qwen2.5 models without any continual training. We observe that QOFT consistently outperforms both QLoRA and the base model across all evaluated scales and tasks, despite using significantly fewer trainable parameters. For instance, on the Qwen2.5-7B instruction-tuned model, QOFT achieves a 96.9\% SAT Math accuracy compared to QLoRA’s 68.8\%, while utilizing only 17.55M parameters (57\% fewer than QLoRA’s 40.37M). This advantage scales robustly: the Qwen2.5-32B variant finetuned with QOFT attains 100\% SAT Math accuracy, surpassing both the baseline (65.6\%) and QLoRA (96.9\%). These gains persist across mathematical reasoning tasks (e.g., 70.0\% on AMC23 for QOFT-32B vs. QLoRA’s 40.0\%), suggesting that orthogonal adaptation in quantized space better preserves the model’s reasoning capabilities compared to low-rank adaptation. The results demonstrate QOFT’s dual strength: parameter efficiency without sacrificing task performance, particularly in the quantized setting. In contrast, QLoRA-finetuned models can exhibit 
training instabilities~\cite{li2023loftqlorafinetuningawarequantizationlarge}, leading to model collapse where their performance fell below the base model. Appendix~\ref{app:math} gives more results on finetuning math-specific Qwen2.5 models.

\subsection{Text-to-image Generative Models: SD-3.5}
To assay the generality of the proposed methods across modalities, we perform Dreambooth~\cite{ruiz2023dreambooth} finetuning on the latest Stable Diffusion 3.5 models~\cite{esser2024scalingrectifiedflowtransformers}. Dreambooth finetunes text-to-image models using a limited set of images depicting the same subject. This process binds the subject to a unique token identifier, enabling subject-driven generation where the model synthesizes this subject in
novel scenes beyond the training data. Qualitative results are shown in Figure~\ref{fig:sd3_large} and Appendix~\ref{app:sd}. We also report the actual peak GPU memory usage during the finetuning process in Appendix~\ref{app:sd}. For finetuning the NormalFloat 4-bit quantized Stable Diffusion 3.5 Large model, QOFT requires slightly less GPU memory ($35.02$ GB) than the QLoRA method ($35.03$ GiB).

\section{Concluding Remarks}

OFTv2 advances orthogonal finetuning through three key innovations: (i) an input-centric reformulation using matrix–vector products, reducing training time by over 10× and peak memory by ~3× without loss in performance; (ii) a Neumann series based approximation of the Cayley transform, improving numerical stability while preserving approximate orthogonality; and (iii) an extension to quantized models, which matches or surpasses QLoRA in speed, stability, and memory efficiency. Across BART, LLaMA2, Qwen2.5, and Stable Diffusion3.5 (0.5B–72B), OFTv2 achieves competitive performance with roughly half the trainable parameters and consistent memory savings.

\section{Limitations}
OFTv2 substantially improves upon OFT in both memory and computational efficiency, matching low-rank methods in memory usage across data types and training speed in the quantized setting. However, its full-precision fine-tuning remains slower. This limitation arises from fundamental differences: low-rank can be naturally maintained efficiently through two simple linear layers, while preserving orthogonality presents a greater optimization challenge. Additionally, low-rank approaches benefit from extensive community-driven engineering and optimization. Bridging this computational gap presents an interesting research direction.

\section*{Acknowledgment}
The authors would like to sincerely thank Tim Z. Xiao, Le Chen, Yao Feng and Zhen Liu for suggestions and helpful discussions. The core idea was proposed by WL and ZQ, the experiments were conducted by ZQ, and the project was led and supervised by WL. The paper was drafted by WL and ZQ, and later polished by AW and BS.

\bibliography{ref}

\begin{thebibliography}{77}
\providecommand{\natexlab}[1]{#1}

\bibitem[{Aghajanyan et~al.(2020)Aghajanyan, Zettlemoyer, and
  Gupta}]{aghajanyan2020intrinsic}
Armen Aghajanyan, Luke Zettlemoyer, and Sonal Gupta. 2020.
\newblock Intrinsic dimensionality explains the effectiveness of language model
  fine-tuning.
\newblock \emph{arXiv preprint arXiv:2012.13255}.

\bibitem[{Ansell et~al.(2022)Ansell, Ponti, Korhonen, and
  Vuli{\'c}}]{ansell2022composable}
Alan Ansell, Edoardo Ponti, Anna Korhonen, and Ivan Vuli{\'c}. 2022.
\newblock Composable sparse fine-tuning for cross-lingual transfer.
\newblock In \emph{ACL}.

\bibitem[{Ashkboos et~al.(2024)Ashkboos, Mohtashami, Croci, Li, Cameron, Jaggi,
  Alistarh, Hoefler, and Hensman}]{ashkboos2024quarot}
Saleh Ashkboos, Amirkeivan Mohtashami, Maximilian Croci, Bo~Li, Pashmina
  Cameron, Martin Jaggi, Dan Alistarh, Torsten Hoefler, and James Hensman.
  2024.
\newblock Quarot: Outlier-free 4-bit inference in rotated llms.
\newblock In \emph{NeurIPS}.

\bibitem[{Bini et~al.(2024)Bini, Roth, Akata, and Khoreva}]{bini2024ether}
Massimo Bini, Karsten Roth, Zeynep Akata, and Anna Khoreva. 2024.
\newblock Ether: Efficient finetuning of large-scale models with hyperplane
  reflections.
\newblock In \emph{ICML}.

\bibitem[{Chavan et~al.(2023)Chavan, Liu, Gupta, Xing, and
  Shen}]{chavan2023one}
Arnav Chavan, Zhuang Liu, Deepak Gupta, Eric Xing, and Zhiqiang Shen. 2023.
\newblock One-for-all: Generalized lora for parameter-efficient fine-tuning.
\newblock \emph{arXiv preprint arXiv:2306.07967}.

\bibitem[{Chen et~al.(2023)Chen, Zhang, Shi, Li, Smola, and
  Yang}]{chen2022parameter}
Jiaao Chen, Aston Zhang, Xingjian Shi, Mu~Li, Alex Smola, and Diyi Yang. 2023.
\newblock Parameter-efficient fine-tuning design spaces.
\newblock In \emph{ICLR}.

\bibitem[{Chen(2005)}]{chen2005matrix}
Ke~Chen. 2005.
\newblock \emph{Matrix preconditioning techniques and applications}.
\newblock 19. Cambridge University Press.

\bibitem[{Chen et~al.(2022)Chen, Ge, Tong, Wang, Song, Wang, and
  Luo}]{chen2022adaptformer}
Shoufa Chen, Chongjian Ge, Zhan Tong, Jiangliu Wang, Yibing Song, Jue Wang, and
  Ping Luo. 2022.
\newblock Adaptformer: Adapting vision transformers for scalable visual
  recognition.
\newblock In \emph{NeurIPS}.

\bibitem[{Cobbe et~al.(2021)Cobbe, Kosaraju, Bavarian, Chen, Jun, Kaiser,
  Plappert, Tworek, Hilton, Nakano et~al.}]{cobbe2021training}
Karl Cobbe, Vineet Kosaraju, Mohammad Bavarian, Mark Chen, Heewoo Jun, Lukasz
  Kaiser, Matthias Plappert, Jerry Tworek, Jacob Hilton, Reiichiro Nakano, and
  1 others. 2021.
\newblock Training verifiers to solve math word problems.
\newblock \emph{arXiv preprint arXiv:2110.14168}.

\bibitem[{Dettmers et~al.(2023)Dettmers, Pagnoni, Holtzman, and
  Zettlemoyer}]{dettmers2023qlora}
Tim Dettmers, Artidoro Pagnoni, Ari Holtzman, and Luke Zettlemoyer. 2023.
\newblock Qlora: Efficient finetuning of quantized llms.
\newblock In \emph{NeurIPS}.

\bibitem[{Edalati et~al.(2022)Edalati, Tahaei, Kobyzev, Nia, Clark, and
  Rezagholizadeh}]{edalati2022krona}
Ali Edalati, Marzieh Tahaei, Ivan Kobyzev, Vahid~Partovi Nia, James~J Clark,
  and Mehdi Rezagholizadeh. 2022.
\newblock Krona: Parameter efficient tuning with kronecker adapter.
\newblock \emph{arXiv preprint arXiv:2212.10650}.

\bibitem[{Esser et~al.(2024)Esser, Kulal, Blattmann, Entezari, M{\"u}ller,
  Saini, Levi, Lorenz, Sauer, Boesel
  et~al.}]{esser2024scalingrectifiedflowtransformers}
Patrick Esser, Sumith Kulal, Andreas Blattmann, Rahim Entezari, Jonas
  M{\"u}ller, Harry Saini, Yam Levi, Dominik Lorenz, Axel Sauer, Frederic
  Boesel, and 1 others. 2024.
\newblock Scaling rectified flow transformers for high-resolution image
  synthesis.
\newblock In \emph{ICML}.

\bibitem[{Feng et~al.(2025)Feng, Pu, Hu, Li, Ai, and Wang}]{feng2025omoe}
Jinyuan Feng, Zhiqiang Pu, Tianyi Hu, Dongmin Li, Xiaolin Ai, and Huimu Wang.
  2025.
\newblock Omoe: Diversifying mixture of low-rank adaptation by orthogonal
  finetuning.
\newblock \emph{arXiv preprint arXiv:2501.10062}.

\bibitem[{Gheini et~al.(2021)Gheini, Ren, and May}]{gheini2021cross}
Mozhdeh Gheini, Xiang Ren, and Jonathan May. 2021.
\newblock Cross-attention is all you need: Adapting pretrained transformers for
  machine translation.
\newblock In \emph{EMNLP}.

\bibitem[{Gorbunov et~al.(2024)Gorbunov, Yudin, Soboleva, Alanov, Naumov, and
  Rakhuba}]{gorbunov2024group}
Mikhail Gorbunov, Kolya Yudin, Vera Soboleva, Aibek Alanov, Alexey Naumov, and
  Maxim Rakhuba. 2024.
\newblock Group and shuffle: Efficient structured orthogonal parametrization.
\newblock In \emph{NeurIPS}.

\bibitem[{Grattafiori et~al.(2024)Grattafiori, Dubey, Jauhri, Pandey, Kadian,
  Al-Dahle, Letman, Mathur, Schelten, Vaughan et~al.}]{grattafiori2024llama}
Aaron Grattafiori, Abhimanyu Dubey, Abhinav Jauhri, Abhinav Pandey, Abhishek
  Kadian, Ahmad Al-Dahle, Aiesha Letman, Akhil Mathur, Alan Schelten, Alex
  Vaughan, and 1 others. 2024.
\newblock The llama 3 herd of models.
\newblock \emph{arXiv preprint arXiv:2407.21783}.

\bibitem[{Guo et~al.(2025)Guo, Yang, Zhang, Song, Zhang, Xu, Zhu, Ma, Wang, Bi
  et~al.}]{deepseekai2025deepseekr1incentivizingreasoningcapability}
Daya Guo, Dejian Yang, Haowei Zhang, Junxiao Song, Ruoyu Zhang, Runxin Xu,
  Qihao Zhu, Shirong Ma, Peiyi Wang, Xiao Bi, and 1 others. 2025.
\newblock Deepseek-r1: Incentivizing reasoning capability in llms via
  reinforcement learning.
\newblock \emph{arXiv preprint arXiv:2501.12948}.

\bibitem[{Guo et~al.(2020)Guo, Rush, and Kim}]{guo2020parameter}
Demi Guo, Alexander~M Rush, and Yoon Kim. 2020.
\newblock Parameter-efficient transfer learning with diff pruning.
\newblock \emph{arXiv preprint arXiv:2012.07463}.

\bibitem[{He et~al.(2024)He, Luo, Bai, Hu, Thai, Shen, Hu, Han, Huang, Zhang
  et~al.}]{he2024olympiadbenchchallengingbenchmarkpromoting}
Chaoqun He, Renjie Luo, Yuzhuo Bai, Shengding Hu, Zhen~Leng Thai, Junhao Shen,
  Jinyi Hu, Xu~Han, Yujie Huang, Yuxiang Zhang, and 1 others. 2024.
\newblock Olympiadbench: A challenging benchmark for promoting agi with
  olympiad-level bilingual multimodal scientific problems.
\newblock \emph{arXiv preprint arXiv:2402.14008}.

\bibitem[{He et~al.(2021)He, Zhou, Ma, Berg-Kirkpatrick, and
  Neubig}]{he2021towards}
Junxian He, Chunting Zhou, Xuezhe Ma, Taylor Berg-Kirkpatrick, and Graham
  Neubig. 2021.
\newblock Towards a unified view of parameter-efficient transfer learning.
\newblock \emph{arXiv preprint arXiv:2110.04366}.

\bibitem[{Hermann et~al.(2015)Hermann, Kocisky, Grefenstette, Espeholt, Kay,
  Suleyman, and Blunsom}]{hermann2015teachingmachinesreadcomprehend}
Karl~Moritz Hermann, Tomas Kocisky, Edward Grefenstette, Lasse Espeholt, Will
  Kay, Mustafa Suleyman, and Phil Blunsom. 2015.
\newblock Teaching machines to read and comprehend.
\newblock In \emph{NIPS}.

\bibitem[{Houlsby et~al.(2019)Houlsby, Giurgiu, Jastrzebski, Morrone,
  De~Laroussilhe, Gesmundo, Attariyan, and Gelly}]{houlsby2019parameter}
Neil Houlsby, Andrei Giurgiu, Stanislaw Jastrzebski, Bruna Morrone, Quentin
  De~Laroussilhe, Andrea Gesmundo, Mona Attariyan, and Sylvain Gelly. 2019.
\newblock Parameter-efficient transfer learning for nlp.
\newblock In \emph{ICML}.

\bibitem[{Hu et~al.(2022{\natexlab{a}})Hu, Wallis, Allen-Zhu, Li, Wang, Wang,
  Chen et~al.}]{hulora2022}
Edward~J Hu, Phillip Wallis, Zeyuan Allen-Zhu, Yuanzhi Li, Shean Wang, Lu~Wang,
  Weizhu Chen, and 1 others. 2022{\natexlab{a}}.
\newblock Lora: Low-rank adaptation of large language models.
\newblock In \emph{ICLR}.

\bibitem[{Hu et~al.(2022{\natexlab{b}})Hu, yelong shen, Wallis, Allen-Zhu, Li,
  Wang, Wang, and Chen}]{hu2022lora}
Edward~J. Hu, yelong shen, Phillip Wallis, Zeyuan Allen-Zhu, Yuanzhi Li, Shean
  Wang, Lu~Wang, and Weizhu Chen. 2022{\natexlab{b}}.
\newblock Lo{RA}: Low-rank adaptation of large language models.
\newblock In \emph{ICLR}.

\bibitem[{Jia et~al.(2022)Jia, Tang, Chen, Cardie, Belongie, Hariharan, and
  Lim}]{jia2022visual}
Menglin Jia, Luming Tang, Bor-Chun Chen, Claire Cardie, Serge Belongie, Bharath
  Hariharan, and Ser-Nam Lim. 2022.
\newblock Visual prompt tuning.
\newblock In \emph{ECCV}.

\bibitem[{Jie and Deng(2023)}]{jie2023fact}
Shibo Jie and Zhi-Hong Deng. 2023.
\newblock Fact: Factor-tuning for lightweight adaptation on vision transformer.
\newblock In \emph{AAAI}.

\bibitem[{Karimi~Mahabadi et~al.(2021)Karimi~Mahabadi, Henderson, and
  Ruder}]{karimi2021compacter}
Rabeeh Karimi~Mahabadi, James Henderson, and Sebastian Ruder. 2021.
\newblock Compacter: Efficient low-rank hypercomplex adapter layers.
\newblock In \emph{NeurIPS}.

\bibitem[{Kim et~al.(2023)Kim, Lee, Kim, Park, Yoo, Kwon, and
  Lee}]{kim2023memoryefficientfinetuningcompressedlarge}
Jeonghoon Kim, Jung~Hyun Lee, Sungdong Kim, Joonsuk Park, Kang~Min Yoo, Se~Jung
  Kwon, and Dongsoo Lee. 2023.
\newblock Memory-efficient fine-tuning of compressed large language models via
  sub-4-bit integer quantization.
\newblock In \emph{NeurIPS}.

\bibitem[{Kingma and Ba(2015)}]{kingma2014adam}
Diederik~P Kingma and Jimmy Ba. 2015.
\newblock Adam: A method for stochastic optimization.
\newblock In \emph{ICLR}.

\bibitem[{Lester et~al.(2021)Lester, Al-Rfou, and Constant}]{lester2021power}
Brian Lester, Rami Al-Rfou, and Noah Constant. 2021.
\newblock The power of scale for parameter-efficient prompt tuning.
\newblock \emph{arXiv preprint arXiv:2104.08691}.

\bibitem[{Lewis et~al.(2019)Lewis, Liu, Goyal, Ghazvininejad, Mohamed, Levy,
  Stoyanov, and
  Zettlemoyer}]{lewis2019bartdenoisingsequencetosequencepretraining}
Mike Lewis, Yinhan Liu, Naman Goyal, Marjan Ghazvininejad, Abdelrahman Mohamed,
  Omer Levy, Ves Stoyanov, and Luke Zettlemoyer. 2019.
\newblock Bart: Denoising sequence-to-sequence pre-training for natural
  language generation, translation, and comprehension.
\newblock \emph{arXiv preprint arXiv:1910.13461}.

\bibitem[{Lewkowycz et~al.(2022)Lewkowycz, Andreassen, Dohan, Dyer,
  Michalewski, Ramasesh, Slone, Anil, Schlag, Gutman-Solo
  et~al.}]{lewkowycz2022solvingquantitativereasoningproblems}
Aitor Lewkowycz, Anders Andreassen, David Dohan, Ethan Dyer, Henryk
  Michalewski, Vinay Ramasesh, Ambrose Slone, Cem Anil, Imanol Schlag, Theo
  Gutman-Solo, and 1 others. 2022.
\newblock Solving quantitative reasoning problems with language models.
\newblock In \emph{NeurIPS}.

\bibitem[{Li and Liang(2021)}]{li2021prefix}
Xiang~Lisa Li and Percy Liang. 2021.
\newblock Prefix-tuning: Optimizing continuous prompts for generation.
\newblock In \emph{ACL}.

\bibitem[{Li et~al.(2023)Li, Yu, Liang, He, Karampatziakis, Chen, and
  Zhao}]{li2023loftqlorafinetuningawarequantizationlarge}
Yixiao Li, Yifan Yu, Chen Liang, Pengcheng He, Nikos Karampatziakis, Weizhu
  Chen, and Tuo Zhao. 2023.
\newblock Loftq: Lora-fine-tuning-aware quantization for large language models.
\newblock \emph{arXiv preprint arXiv:2310.08659}.

\bibitem[{Lian et~al.(2022)Lian, Zhou, Feng, and Wang}]{lian2022scaling}
Dongze Lian, Daquan Zhou, Jiashi Feng, and Xinchao Wang. 2022.
\newblock Scaling \& shifting your features: A new baseline for efficient model
  tuning.
\newblock In \emph{NeurIPS}.

\bibitem[{Liao and Monz(2024)}]{liao20243}
Baohao Liao and Christof Monz. 2024.
\newblock 3-in-1: 2d rotary adaptation for efficient finetuning, efficient
  batching and composability.
\newblock \emph{arXiv preprint arXiv:2409.00119}.

\bibitem[{Liao et~al.(2024)Liao, Luo, Li, Wu, and
  Fan}]{liao2024mariomathreasoningcode}
Minpeng Liao, Wei Luo, Chengxi Li, Jing Wu, and Kai Fan. 2024.
\newblock Mario: Math reasoning with code interpreter output--a reproducible
  pipeline.
\newblock \emph{arXiv preprint arXiv:2401.08190}.

\bibitem[{Lin et~al.(2024)Lin, Tang, Tang, Yang, Chen, Wang, Xiao, Dang, Gan,
  and Han}]{lin2024awqactivationawareweightquantization}
Ji~Lin, Jiaming Tang, Haotian Tang, Shang Yang, Wei-Ming Chen, Wei-Chen Wang,
  Guangxuan Xiao, Xingyu Dang, Chuang Gan, and Song Han. 2024.
\newblock Awq: Activation-aware weight quantization for on-device llm
  compression and acceleration.
\newblock In \emph{MLSys}.

\bibitem[{Lin et~al.(2020)Lin, Liu, Liu, Feng, Yu, Rehg, Xiong, and
  Song}]{lin2020regularizing}
Rongmei Lin, Weiyang Liu, Zhen Liu, Chen Feng, Zhiding Yu, James~M Rehg,
  Li~Xiong, and Le~Song. 2020.
\newblock Regularizing neural networks via minimizing hyperspherical energy.
\newblock In \emph{CVPR}.

\bibitem[{Ling et~al.(2017)Ling, Yogatama, Dyer, and
  Blunsom}]{ling2017programinductionrationalegeneration}
Wang Ling, Dani Yogatama, Chris Dyer, and Phil Blunsom. 2017.
\newblock Program induction by rationale generation: Learning to solve and
  explain algebraic word problems.
\newblock \emph{arXiv preprint arXiv:1705.04146}.

\bibitem[{Lingam et~al.(2024)Lingam, Neerkaje, Vavre, Shetty, Gudur, Ghosh,
  Choi, Dimakis, Bojchevski, and Sanghavi}]{lingam2024svft}
Vijay~Chandra Lingam, Atula Neerkaje, Aditya Vavre, Aneesh Shetty,
  Gautham~Krishna Gudur, Joydeep Ghosh, Eunsol Choi, Alex Dimakis, Aleksandar
  Bojchevski, and Sujay Sanghavi. 2024.
\newblock Svft: Parameter-efficient fine-tuning with singular vectors.
\newblock In \emph{NeurIPS}.

\bibitem[{Liu et~al.(2022)Liu, Tam, Muqeeth, Mohta, Huang, Bansal, and
  Raffel}]{liu2022few}
Haokun Liu, Derek Tam, Mohammed Muqeeth, Jay Mohta, Tenghao Huang, Mohit
  Bansal, and Colin~A Raffel. 2022.
\newblock Few-shot parameter-efficient fine-tuning is better and cheaper than
  in-context learning.
\newblock In \emph{NeurIPS}.

\bibitem[{Liu et~al.(2018)Liu, Lin, Liu, Liu, Yu, Dai, and
  Song}]{liu2018learning}
Weiyang Liu, Rongmei Lin, Zhen Liu, Lixin Liu, Zhiding Yu, Bo~Dai, and Le~Song.
  2018.
\newblock Learning towards minimum hyperspherical energy.
\newblock In \emph{NeurIPS}.

\bibitem[{Liu et~al.(2021{\natexlab{a}})Liu, Lin, Liu, Rehg, Paull, Xiong,
  Song, and Weller}]{liu2021orthogonal}
Weiyang Liu, Rongmei Lin, Zhen Liu, James~M Rehg, Liam Paull, Li~Xiong,
  Le~Song, and Adrian Weller. 2021{\natexlab{a}}.
\newblock Orthogonal over-parameterized training.
\newblock In \emph{CVPR}.

\bibitem[{Liu et~al.(2021{\natexlab{b}})Liu, Lin, Liu, Xiong, Sch{\"o}lkopf,
  and Weller}]{liu2021learning}
Weiyang Liu, Rongmei Lin, Zhen Liu, Li~Xiong, Bernhard Sch{\"o}lkopf, and
  Adrian Weller. 2021{\natexlab{b}}.
\newblock Learning with hyperspherical uniformity.
\newblock In \emph{AISTATS}.

\bibitem[{Liu et~al.(2024)Liu, Qiu, Feng, Xiu, Xue, Yu, Feng, Liu, Heo, Peng,
  Wen, Black, Weller, and Sch{\"o}lkopf}]{liu2024boft}
Weiyang Liu, Zeju Qiu, Yao Feng, Yuliang Xiu, Yuxuan Xue, Longhui Yu, Haiwen
  Feng, Zhen Liu, Juyeon Heo, Songyou Peng, Yandong Wen, Michael~J. Black,
  Adrian Weller, and Bernhard Sch{\"o}lkopf. 2024.
\newblock Parameter-efficient orthogonal finetuning via butterfly
  factorization.
\newblock In \emph{ICLR}.

\bibitem[{Liu et~al.(2023)Liu, Yu, Weller, and
  Sch{\"o}lkopf}]{liu2023generalizing}
Weiyang Liu, Longhui Yu, Adrian Weller, and Bernhard Sch{\"o}lkopf. 2023.
\newblock Generalizing and decoupling neural collapse via hyperspherical
  uniformity gap.
\newblock In \emph{ICLR}.

\bibitem[{Luo et~al.(2023)Luo, Huang, Zhou, Sun, Jiang, Wang, and
  Ji}]{luo2023towards}
Gen Luo, Minglang Huang, Yiyi Zhou, Xiaoshuai Sun, Guannan Jiang, Zhiyu Wang,
  and Rongrong Ji. 2023.
\newblock Towards efficient visual adaption via structural re-parameterization.
\newblock \emph{arXiv preprint arXiv:2302.08106}.

\bibitem[{Ma et~al.(2024)Ma, Chu, Yang, Lin, Gao, and Zhao}]{ma2024parameter}
Xinyu Ma, Xu~Chu, Zhibang Yang, Yang Lin, Xin Gao, and Junfeng Zhao. 2024.
\newblock Parameter efficient quasi-orthogonal fine-tuning via givens rotation.
\newblock In \emph{ICML}.

\bibitem[{Mao et~al.(2021)Mao, Mathias, Hou, Almahairi, Ma, Han, Yih, and
  Khabsa}]{mao2021unipelt}
Yuning Mao, Lambert Mathias, Rui Hou, Amjad Almahairi, Hao Ma, Jiawei Han,
  Wen-tau Yih, and Madian Khabsa. 2021.
\newblock Unipelt: A unified framework for parameter-efficient language model
  tuning.
\newblock \emph{arXiv preprint arXiv:2110.07577}.

\bibitem[{Merity et~al.(2017)Merity, Xiong, Bradbury, and
  Socher}]{merity2016pointersentinelmixturemodels}
Stephen Merity, Caiming Xiong, James Bradbury, and Richard Socher. 2017.
\newblock Pointer sentinel mixture models.
\newblock In \emph{ICLR}.

\bibitem[{Narayan et~al.(2018)Narayan, Cohen, and
  Lapata}]{narayan2018dontdetailsjustsummary}
Shashi Narayan, Shay~B Cohen, and Mirella Lapata. 2018.
\newblock Don't give me the details, just the summary! topic-aware
  convolutional neural networks for extreme summarization.
\newblock \emph{arXiv preprint arXiv:1808.08745}.

\bibitem[{OpenR1-Team(2025)}]{open-r1}
OpenR1-Team. 2025.
\newblock \href {https://huggingface.co/datasets/open-r1/OpenR1-Math-220k}
  {Openr1-math-220k}.

\bibitem[{Project-Numina()}]{amc23}
Project-Numina.
\newblock \href {https://huggingface.co/datasets/AI-MO/aimo-validation-amc}
  {Aimo validation amc}.

\bibitem[{Qiu et~al.(2025)Qiu, Buchholz, Xiao, Dax, Sch\"olkopf, and
  Liu}]{qiu2025poet}
Zeju Qiu, Simon Buchholz, Tim~Z. Xiao, Maximilian Dax, Bernhard Sch\"olkopf,
  and Weiyang Liu. 2025.
\newblock Reparameterized llm training via orthogonal equivalence
  transformation.
\newblock \emph{arXiv preprint arXiv:2506.08001}.

\bibitem[{Qiu et~al.(2023)Qiu, Liu, Feng, Xue, Feng, Liu, Zhang, Weller, and
  Sch{\"o}lkopf}]{qiu2023controlling}
Zeju Qiu, Weiyang Liu, Haiwen Feng, Yuxuan Xue, Yao Feng, Zhen Liu, Dan Zhang,
  Adrian Weller, and Bernhard Sch{\"o}lkopf. 2023.
\newblock Controlling text-to-image diffusion by orthogonal finetuning.
\newblock In \emph{NeurIPS}.

\bibitem[{Raj and Coyle(2025)}]{raj2025hyper}
Snehal Raj and Brian Coyle. 2025.
\newblock Hyper compressed fine-tuning of large foundation models with quantum
  inspired adapters.
\newblock \emph{arXiv preprint arXiv:2502.06916}.

\bibitem[{Ruiz et~al.(2023)Ruiz, Li, Jampani, Pritch, Rubinstein, and
  Aberman}]{ruiz2023dreambooth}
Nataniel Ruiz, Yuanzhen Li, Varun Jampani, Yael Pritch, Michael Rubinstein, and
  Kfir Aberman. 2023.
\newblock Dreambooth: Fine tuning text-to-image diffusion models for
  subject-driven generation.
\newblock In \emph{CVPR}.

\bibitem[{Su et~al.(2024)Su, Liu, Qiu, Liu, and Xu}]{su2024defense}
Junda Su, Zirui Liu, Zeju Qiu, Weiyang Liu, and Zhaozhuo Xu. 2024.
\newblock In defense of structural sparse adapters for concurrent llm serving.
\newblock In \emph{Findings of EMNLP}.

\bibitem[{Sung et~al.(2022)Sung, Cho, and Bansal}]{sung2022lst}
Yi-Lin Sung, Jaemin Cho, and Mohit Bansal. 2022.
\newblock Lst: Ladder side-tuning for parameter and memory efficient transfer
  learning.
\newblock In \emph{NeurIPS}.

\bibitem[{Sung et~al.(2021)Sung, Nair, and Raffel}]{sung2021training}
Yi-Lin Sung, Varun Nair, and Colin~A Raffel. 2021.
\newblock Training neural networks with fixed sparse masks.
\newblock \emph{NeurIPS}.

\bibitem[{Tseng et~al.(2024)Tseng, Chee, Sun, Kuleshov, and
  De~Sa}]{tseng2024quip}
Albert Tseng, Jerry Chee, Qingyao Sun, Volodymyr Kuleshov, and Christopher
  De~Sa. 2024.
\newblock Quip\#: Even better llm quantization with hadamard incoherence and
  lattice codebooks.
\newblock \emph{arXiv preprint arXiv:2402.04396}.

\bibitem[{Valipour et~al.(2022)Valipour, Rezagholizadeh, Kobyzev, and
  Ghodsi}]{valipour2022dylora}
Mojtaba Valipour, Mehdi Rezagholizadeh, Ivan Kobyzev, and Ali Ghodsi. 2022.
\newblock Dylora: Parameter efficient tuning of pre-trained models using
  dynamic search-free low-rank adaptation.
\newblock \emph{arXiv preprint arXiv:2210.07558}.

\bibitem[{Vu et~al.(2022)Vu, Lester, Constant, Al-Rfou, and Cer}]{vu2022spot}
Tu~Vu, Brian Lester, Noah Constant, Rami Al-Rfou, and Daniel Cer. 2022.
\newblock Spot: Better frozen model adaptation through soft prompt transfer.
\newblock In \emph{ACL}.

\bibitem[{Wang et~al.(2022)Wang, Mukherjee, Liu, Gao, Awadallah, and
  Gao}]{wang2022adamix}
Yaqing Wang, Subhabrata Mukherjee, Xiaodong Liu, Jing Gao, Ahmed~Hassan
  Awadallah, and Jianfeng Gao. 2022.
\newblock Adamix: Mixture-of-adapter for parameter-efficient tuning of large
  language models.
\newblock In \emph{EMNLP}.

\bibitem[{Wei et~al.(2023)Wei, Luan, Liu, Dong, and
  Wang}]{wei2023cmathlanguagemodelpass}
Tianwen Wei, Jian Luan, Wei Liu, Shuang Dong, and Bin Wang. 2023.
\newblock Cmath: Can your language model pass chinese elementary school math
  test?
\newblock \emph{arXiv preprint arXiv:2306.16636}.

\bibitem[{Wu et~al.(2024)Wu, Wang, Zhao, and Wong}]{wu2024mixture}
Taiqiang Wu, Jiahao Wang, Zhe Zhao, and Ngai Wong. 2024.
\newblock Mixture-of-subspaces in low-rank adaptation.
\newblock \emph{arXiv preprint arXiv:2406.11909}.

\bibitem[{Yang et~al.(2024{\natexlab{a}})Yang, Yang, Zhang, Hui, Zheng, Yu, Li,
  Liu, Huang, Wei et~al.}]{yang2024qwen2}
An~Yang, Baosong Yang, Beichen Zhang, Binyuan Hui, Bo~Zheng, Bowen Yu,
  Chengyuan Li, Dayiheng Liu, Fei Huang, Haoran Wei, and 1 others.
  2024{\natexlab{a}}.
\newblock Qwen2.5 technical report.
\newblock \emph{arXiv preprint arXiv:2412.15115}.

\bibitem[{Yang et~al.(2024{\natexlab{b}})Yang, Jia, Gu, Lin, Chen, Pang, Yin,
  Sun, Wu, and Wang}]{yang2024orthogonal}
Chenxu Yang, Ruipeng Jia, Naibin Gu, Zheng Lin, Siyuan Chen, Chao Pang,
  Weichong Yin, Yu~Sun, Hua Wu, and Weiping Wang. 2024{\natexlab{b}}.
\newblock Orthogonal finetuning for direct preference optimization.
\newblock \emph{arXiv preprint arXiv:2409.14836}.

\bibitem[{Yuan et~al.(2024)Yuan, Liu, and Xu}]{yuan2024bridging}
Shen Yuan, Haotian Liu, and Hongteng Xu. 2024.
\newblock Bridging the gap between low-rank and orthogonal adaptation via
  householder reflection adaptation.
\newblock In \emph{NeurIPS}.

\bibitem[{Zaken et~al.(2022)Zaken, Goldberg, and Ravfogel}]{zaken2022bitfit}
Elad~Ben Zaken, Yoav Goldberg, and Shauli Ravfogel. 2022.
\newblock {BitFit: Simple Parameter-efficient Fine-tuning for Transformer-based
  Masked Language-models}.
\newblock In \emph{ACL}.

\bibitem[{Zhang et~al.(2023{\natexlab{a}})Zhang, Li, Chen, Jiang, Wang, and
  Qian}]{zhang2023increlora}
Feiyu Zhang, Liangzhi Li, Junhao Chen, Zhouqiang Jiang, Bowen Wang, and Yiming
  Qian. 2023{\natexlab{a}}.
\newblock Increlora: Incremental parameter allocation method for
  parameter-efficient fine-tuning.
\newblock \emph{arXiv preprint arXiv:2308.12043}.

\bibitem[{Zhang et~al.(2023{\natexlab{b}})Zhang, Chen, Bukharin, He, Cheng,
  Chen, and Zhao}]{zhang2022adaptive}
Qingru Zhang, Minshuo Chen, Alexander Bukharin, Pengcheng He, Yu~Cheng, Weizhu
  Chen, and Tuo Zhao. 2023{\natexlab{b}}.
\newblock Adaptive budget allocation for parameter-efficient fine-tuning.
\newblock In \emph{ICLR}.

\bibitem[{Zhang et~al.(2023{\natexlab{c}})Zhang, Chen, Bukharin,
  Karampatziakis, He, Cheng, Chen, and Zhao}]{zhang2023adalora}
Qingru Zhang, Minshuo Chen, Alexander Bukharin, Nikos Karampatziakis, Pengcheng
  He, Yu~Cheng, Weizhu Chen, and Tuo Zhao. 2023{\natexlab{c}}.
\newblock Adalora: Adaptive budget allocation for parameter-efficient
  fine-tuning.
\newblock \emph{arXiv preprint arXiv:2303.10512}.

\bibitem[{Zhang et~al.(2024)Zhang, Qiang, Somayajula, and
  Xie}]{zhang2024autolora}
Ruiyi Zhang, Rushi Qiang, Sai~Ashish Somayajula, and Pengtao Xie. 2024.
\newblock Autolora: Automatically tuning matrix ranks in low-rank adaptation
  based on meta learning.
\newblock \emph{arXiv preprint arXiv:2403.09113}.

\bibitem[{Zhang et~al.(2022)Zhang, Zhou, and Liu}]{zhang2022neural}
Yuanhan Zhang, Kaiyang Zhou, and Ziwei Liu. 2022.
\newblock Neural prompt search.
\newblock \emph{arXiv preprint arXiv:2206.04673}.

\bibitem[{Zi et~al.(2023)Zi, Qi, Wang, Wang, Wong, and Zhang}]{zi2023delta}
Bojia Zi, Xianbiao Qi, Lingzhi Wang, Jianan Wang, Kam-Fai Wong, and Lei Zhang.
  2023.
\newblock Delta-lora: Fine-tuning high-rank parameters with the delta of
  low-rank matrices.
\newblock \emph{arXiv preprint arXiv:2309.02411}.

\end{thebibliography}

\newpage
\onecolumn

\addcontentsline{toc}{section}{Appendix} %
\renewcommand \thepart{} %
\renewcommand \partname{}
\part{\Large{\centerline{Appendix}}}
\parttoc

\newpage

\newpage

\appendix

\newpage
\clearpage

\section{Experimental Details}

\label{sec:appendix}

This section outlines the specifics of our experimental setup, including the optimizer, code frameworks, computational resources, evaluation methods, and detailed hyperparameters used for each experiment.

\paragraph{Training details.} We employed the Adam optimizer~\cite{kingma2014adam} for all our training runs. The specific hyperparameters used for each experiment are detailed in the tables referenced below. These include learning rates, batch sizes, number of training epochs, and method-specific configurations: the rank $r$ for LoRA-based methods and the block size $b$ for OFTv2/QOFT. If not explicitly specified, the $r$ for LoRA-based methods is 16 and the block size $b$ for OFTv2/QOFT is set as 32. For the Wikitext dataset, hyperparameters are listed in Table~\ref{tab:hyper_wikitext}. For the GSM8K dataset, hyperparameters are listed in Table~\ref{tab:hyper_gsm8k}. For the XSum dataset, hyperparameters are listed in Table~\ref{tab:hyper_xsum}. For the CNN/DailyMail dataset, hyperparameters are listed in Table~\ref{tab:hyper_cnn}. Since it is known that merging QLoRA adapter weights to its quantized base models leads to performance degradation\footnote{Comparison of merging methods: \url{https://kaitchup.substack.com/p/lora-adapters-when-a-naive-merge}} and distorts the real performance, for every experiment, we evaluate the fine-tuned model without merging the trainable parameters, but load them as extra adapter layers. 

\begin{table*}[h!]
\small
\setlength{\tabcolsep}{2.7pt}
\renewcommand{\arraystretch}{1.35}
\centering
\begin{tabular}{c|cccccc|cccccc}
\multirow{3}*{\bf Hyperparameter} & \multicolumn{6}{c|}{\bf LoRA} & \multicolumn{6}{c}{\bf OFTv2} \\
~ & \multicolumn{3}{c}{BF16} & \multicolumn{3}{c|}{NF4} & \multicolumn{3}{c}{BF16} & \multicolumn{3}{c}{NF4} \\
~ & $r=8$ & $r=16$ & $r=32$ & $r=8$ & $r=16$ & $r=32$ & $b=16$ & $b=32$ & $b=64$ & $b=16$ & $b=32$ & $b=64$ \\
\shline
Learning rate & 1e-4 & 1e-4 & 1e-4 & 1e-4 & 1e-4 & 1e-4 & 4e-4 & 4e-4 & 4e-4 & 4e-4 & 4e-4 & 4e-4  \\ 
Epoch & 10 & 10 & 10 & 10 & 10 & 10 & 5 & 5 & 5 & 5 & 5 & 5  \\ 
Batch size & 32 & 32 & 32 & 32 & 32 & 32 & 32 & 32 & 32 & 32 & 32 & 32  \\ 
Gradient Accumulation & 4 & 4 & 4 & 4 & 4 & 4 & 4 & 4 & 4 & 4 & 4 & 4  \\
\end{tabular}
\caption{Hyper-parameter setup of fine-tuning BART-large on XSum with LoRA and OFTv2.}
\label{tab:hyper_xsum}
\end{table*}

\begin{table*}[h!]
\small
\setlength{\tabcolsep}{2.7pt}
\renewcommand{\arraystretch}{1.35}
\centering
\begin{tabular}{c|cccccc|cccccc}
\multirow{3}*{\bf Hyperparameter} & \multicolumn{6}{c|}{\bf LoRA} & \multicolumn{6}{c}{\bf OFTv2} \\ 
~ & \multicolumn{3}{c}{BF16} & \multicolumn{3}{c|}{NF4} & \multicolumn{3}{c}{BF16} & \multicolumn{3}{c}{NF4} \\
~ & $r=8$ & $r=16$ & $r=32$ & $r=8$ & $r=16$ & $r=32$ & $b=16$ & $b=32$ & $b=64$ & $b=16$ & $b=32$ & $b=64$ \\\shline
Learning rate & 1e-4 & 1e-4 & 1e-4 & 1e-4 & 1e-4 & 1e-4 & 4e-4 & 4e-4 & 4e-4 & 4e-4 & 4e-4 & 4e-4  \\
Epoch & 5 & 5 & 5 & 5 & 5 & 5 & 5 & 5 & 5 & 5 & 5 & 5  \\
Batch size & 64 & 64 & 64 & 64 & 64 & 64 & 64 & 64 & 64 & 64 & 64 & 64  \\
Gradient Accumulation & 4 & 4 & 4 & 4 & 4 & 4 & 4 & 4 & 4 & 4 & 4 & 4  \\
\end{tabular}
\caption{Hyper-parameter setup of fine-tuning BART-large on CNN/DailyMail with LoRA and OFTv2.}
\label{tab:hyper_cnn}
\end{table*}

\begin{table*}[t!]
\small
\setlength{\tabcolsep}{7pt}
\renewcommand{\arraystretch}{1.35}
\begin{center}
\begin{tabular}{c|cccc|cccc}
\multirow{3}*{\bf Hyperparameter} & \multicolumn{4}{c|}{\bf LoRA} & \multicolumn{4}{c}{\bf OFTv2} \\
~ & \multicolumn{2}{c}{BF16} & \multicolumn{2}{c|}{NF4} & \multicolumn{2}{c}{BF16} & \multicolumn{2}{c}{NF4} \\
~ & 7B & 13B & 7B & 13B & 7B & 13B & 7B & 13B \\
\shline
Learning rate & 2e-4 & 2e-4 & 2e-4 & 2e-4 & 2e-4 & 2e-4 & 2e-4 & 2e-4 \\
Epoch & 10 & 10 & 10 & 10 & 10 & 10 & 10 & 10  \\
Batch size & 16 & 16 & 16 & 16 & 16 & 16 & 16 & 16  \\
Gradient Accumulation & 2 & 2 & 2 & 2 & 2 & 2 & 2 & 2  \\
\end{tabular}
\end{center}
\vspace{-1.5mm}
\caption{Hyper-parameter setup of fine-tuning Llama 2 on Wikitext-2 with LoRA and OFTv2.}
\label{tab:hyper_wikitext}
\vspace{2mm}
\end{table*}

\begin{table*}[t!]
\small
\setlength{\tabcolsep}{7pt}
\renewcommand{\arraystretch}{1.35}
\centering
\begin{tabular}{c|cccc|cccc}
\multirow{3}*{\bf Hyperparameter} & \multicolumn{4}{c}{\bf LoRA} & \multicolumn{4}{c}{\bf OFTv2} \\
~ & \multicolumn{2}{c}{BF16} & \multicolumn{2}{c}{NF4} & \multicolumn{2}{c}{BF16} & \multicolumn{2}{c}{NF4} \\
~ & 7B & 13B & 7B & 13B & 7B & 13B & 7B & 13B \\
\shline
Learning rate & 2e-4 & 2e-4 & 2e-4 & 2e-4 & 8e-4 & 8e-4 & 8e-4 & 8e-4 \\ 
Epoch & 10 & 10 & 10 & 10 & 10 & 10 & 10 & 10  \\
Batch size & 16 & 16 & 16 & 16 & 16 & 16 & 16 & 16  \\
Gradient Accumulation & 4 & 4 & 4 & 4 & 4 & 4 & 4 & 4  \\
\end{tabular}
\vspace{-1.5mm}
\caption{Hyper-parameter setup of fine-tuning Llama 2 on GSM8K with LoRA and OFTv2.}
\label{tab:hyper_gsm8k}
\vspace{2mm}
\end{table*}

\paragraph{Code framework.} Our method is implemented using the \textbf{Hugging Face PEFT}\footnote{\url{https://huggingface.co/docs/peft/en/index}} framework, a widely adopted open-source framework providing state-of-the-art parameter-efficient fine-tuning of pre-trained large language models and diffusion models. The implementation of OFTv2 will be released on Hugging Face PEFT soon, to allow for easy reproduction of our training results. We utilized the Hugging Face TRL library for supervised fine-tuning\footnote{\url{https://github.com/huggingface/trl}}. For the base model quantization, we leveraged bitsandbytes\footnote{\url{https://github.com/bitsandbytes-foundation/bitsandbytes}} for the NormalFloat 4-bit quantization and the QLoRA finetuning, and AutoAWQ\footnote{\url{https://github.com/casper-hansen/AutoAWQ}} for AWQ quantization.

\paragraph{Pretrained models.} Our work utilized several pre-trained large language models. Specifically, we employed models from the Qwen2.5 model series\footnote{\url{https://huggingface.co/collections/Qwen/qwen25-66e81a666513e518adb90d9e}}, which are available under the permissive \textbf{Apache 2.0 license}. We also leveraged the Llama 2 models\footnote{\url{https://huggingface.co/collections/meta-llama/metas-llama2-models-675bfd70e574a62dd0e40541}}, governed by the \textbf{Llama 2 license}. Additionally, for the text summarization tasks, the BART-large model was used, which is also distributed under the \textbf{Apache 2.0 license}. For the text-to-image generation, we utilized the Stable Diffusion 3.5 models, which are under the \textbf{Stability AI Community license}. We have adhered to all respective licensing agreements for these models throughout our work. 

\paragraph{Dataset.} The experiments in this study utilized a diverse range of publicly available datasets to ensure comprehensive evaluation. For finetuning language modeling tasks, we employed the Wikitext-2\footnote{\url{https://huggingface.co/datasets/Salesforce/wikitext}} dataset, which is distributed under the \textbf{CC-BY-SA-3.0 license}. Text summarization performance was assessed by fine-tuning on the CNN / DailyMail Dataset\footnote{\url{https://huggingface.co/datasets/abisee/cnn_dailymail}}, also licensed under \textbf{Apache 2.0}, and the XSum dataset\footnote{\url{https://huggingface.co/datasets/EdinburghNLP/xsum}}, which is available under the \textbf{MIT license}. For finetuning mathematical reasoning capabilities, we used the GSM8K\footnote{\url{https://huggingface.co/datasets/openai/gsm8k}} dataset, available under the \textbf{MIT license}, and the OpenR1-Math-220k\footnote{\url{https://huggingface.co/datasets/open-r1/OpenR1-Math-220k}} dataset, which can be used under the \textbf{Apache 2.0 license}. The Dreambooth dataset\footnote{\url{https://huggingface.co/datasets/google/dreambooth}} for fine-tuning the diffusion models are under the \textbf{cc-by-4.0 license}.

\paragraph{Compute Resources.} All the training tasks are performed on a \textbf{NVIDIA HGX H100 8-GPU System} node with 80GB memory each. We used a single \textbf{NVIDIA H100 NVL} GPU with 94GB memory to benchmark the memory usage.

\clearpage
\newpage
\section{Effect of Neumann Series Terms in Orthogonal Parameterization}\label{app:abl}

OFTv2 employs the Cayley-Neumann parameterization to improve the training efficiency; the number of Neumann series terms becomes a hyperparameter. We conducted an additional ablation study to evaluate the impact of the number of Neumann series terms on finetuning performance for WikiText. The results are reported in Table~\ref{tab:abl_neumann}. We observe that when the number of Neumann terms is too small (\eg, 2), the approximation error to orthogonality slightly degrades performance. For the experiments reported in the main paper, we used five Neumann terms, which we found to be well-suited across all evaluated tasks.

\begin{table}[htbp]
\centering
\begin{tabular}{l|lccccc}
\textbf{Model} & \textbf{Method} & \textbf{2 terms} & \textbf{3 terms} & \textbf{4 terms} & \textbf{5 terms} & \textbf{6 terms} \\
\shline
Llama 2 7B  & OFTv2 & 6.22 & 6.15 & 6.14 & 6.13 & 6.14 \\
Llama 2 13B & OFTv2 & 5.11  & 5.00 & 4.99 & 4.98 & 4.99 \\
\hline
Llama 2 7B  & QOFT  & 5.70  & 5.62 & 5.58 & 5.60 & 5.61 \\
Llama 2 13B & QOFT  & 5.14  & 5.02 & 5.04 & 5.05 & 5.05 \\
\end{tabular}
\caption{Effect of Neumann Series Terms on the Llama-2 Models}
\label{tab:abl_neumann}
\end{table}

\clearpage
\newpage
\section{Mathematical Reasoning with Qwen2.5}\label{app:math}

\paragraph{Training details.} We fine-tuned the Qwen2.5 models using QLoRA or QOFT on a random subset of 50,000 samples from the Huggingface OpenR1-Math-220k dataset~\cite{open-r1}. For each method and benchmark, we selected the best-performing model after trying learning rates of $1 \times 10^{-5}$, $2 \times 10^{-5}$, $5 \times 10^{-5}$, and $1 \times 10^{-4}$. We used a batch size of 16 for the 1.5B models and 8 for the 7B and 32B models, with 2 gradient accumulation steps for all. A cosine learning rate scheduler was employed, with a minimum learning rate set to 10\% of the initial value.

\paragraph{Evaluation details.} For evaluating the Qwen2.5 base models and the QLoRA or QOFT fine-tuned versions, we utilized the same evaluation pipeline as Qwen2.5-Math\footnote{\url{https://github.com/QwenLM/Qwen2.5-Math}}. This framework provides robust tools for parsing and evaluating mathematical expressions and problem-solving steps, ensuring accurate and consistent assessment of model performance on these mathematical benchmarks. More specifically, we report the model's pass@1 performance, \ie, the performance on the first attempt for a given task, obtained by utilizing the Qwen2.5 Chain-of-Though question prompt (Figure~\ref{fig:math-cot-prompt}).

\begin{figure*}[h]
\begin{mdframed}[leftmargin=10pt,rightmargin=10pt]
\begin{tabular}{p{0.95\textwidth}}
\texttt{<|im\_start|>system\textbackslash{}n}\\
\texttt{Please reason step by step, and put your final answer within \textbackslash{}\textbackslash{}boxed\{\{\}\}.}\\
\texttt{<|im\_end|>\textbackslash{}n}\\
\texttt{<|im\_start|>user\textbackslash{}n\{input\}<|im\_end|>\textbackslash{}n}\\
\texttt{<|im\_start|>assistant\textbackslash{}n\{output\}\textbackslash{}n\textbackslash{}n}
\end{tabular}
\end{mdframed}
\vspace{-1.5mm}
\caption{Prompt template used for evaluating Qwen2.5 series models on mathematical reasoning benchmarks.}
\label{fig:math-cot-prompt}
\end{figure*}

\begin{table*}[h]
\small
\setlength{\tabcolsep}{4pt}
\renewcommand{\arraystretch}{1.5}
\centering
\begin{tabular}{l|lcccccccc}
\multirow{2}{*}{\textbf{Model}} & \multirow{2}{*}{\textbf{Method}} & \multirow{2}{*}{\textbf{\# Params}} & \multirow{2}{*}{\textbf{AMC23}} & \multirow{2}{*}{\textbf{AQUA}} & \multirow{2}{*}{\textbf{CMATH}} & \textbf{GaoKao} & \textbf{Minerva} & \textbf{Olympiad/} & \textbf{SAT} \\
 & & & & & & \textbf{2023 En} & \textbf{Math} & \textbf{Bench} & \textbf{Math} \\
\shline
 \multirow{2}{*}{\textbf{Qwen2.5-1.5B-math-it}} & \textbf{QLoRA} & 18.46M & 27.5 & 33.5 & 86.8 & 43.6 & 15.4 & 15.1 & 46.9 \\
 & \cellcolor{Gray}\textbf{QOFT} & \cellcolor{Gray} 7.89M & \cellcolor{Gray} \textbf{45.0} & \cellcolor{Gray} \textbf{70.9} & \cellcolor{Gray} \textbf{87.2} & \cellcolor{Gray} \textbf{60.5} & \cellcolor{Gray} \textbf{25.4} & \cellcolor{Gray} \textbf{32.0} & \cellcolor{Gray} \textbf{93.8} \\\hline
 \multirow{2}{*}{\textbf{Qwen2.5-1.5B-math}} & \textbf{QLoRA} & 18.46M & 25.0 & 31.5 & 49.0 & 36.9 & 10.7 & 12.9 & \textbf{50.0} \\
 & \cellcolor{Gray} \textbf{QOFT} & \cellcolor{Gray} 7.89M & \cellcolor{Gray} \textbf{27.5} & \cellcolor{Gray} \textbf{31.5} & \cellcolor{Gray} \textbf{55.5} & \cellcolor{Gray} \textbf{37.7} & \cellcolor{Gray} \textbf{13.6} & \cellcolor{Gray} \textbf{14.4} & \cellcolor{Gray} 37.5 \\
\hline
 \multirow{2}{*}{\textbf{Qwen2.5-7B-math-it}} & \textbf{QLoRA} & 40.37M & 32.5 & 34.6 & 89.8 & 47.0 & 18.8 & 18.2 & 53.1 \\
 & \cellcolor{Gray} \textbf{QOFT} & \cellcolor{Gray} 17.55M & \cellcolor{Gray} \textbf{52.5} & \cellcolor{Gray} \textbf{76.8} & \cellcolor{Gray} \textbf{92.7} & \cellcolor{Gray} \textbf{66.8} & \cellcolor{Gray} \textbf{35.7} & \cellcolor{Gray} \textbf{41.6} & \cellcolor{Gray} \textbf{93.8} \\\hline
 \multirow{2}{*}{\textbf{Qwen2.5-7B-math}} & \textbf{QLoRA} & 40.37M & \textbf{30.0} & 38.6 & 75.7 & 48.6 & 21.0 & \textbf{20.4} & \textbf{50.0} \\
 & \cellcolor{Gray} \textbf{QOFT} & \cellcolor{Gray} 17.55M & \cellcolor{Gray} \textbf{30.0} & \cellcolor{Gray} \textbf{40.6} & \cellcolor{Gray} \textbf{81.7} & \cellcolor{Gray} \textbf{49.4} & \cellcolor{Gray} \textbf{21.3} & \cellcolor{Gray} \textbf{20.4} & \cellcolor{Gray} \textbf{50.0} \\
\end{tabular}
\vspace{-1.5mm}
\caption{The pass@1 performance of the Qwen2.5 series math-specific large language fine-tuned with QLoRA/QOFT by the Chain-of-Thought reasoning.}
\end{table*}

\newpage
\section{Subject-driven Generation with Stable diffusion 3.5}\label{app:sd}
\begin{figure*}[h]
    \centering
    \includegraphics[width=\textwidth]{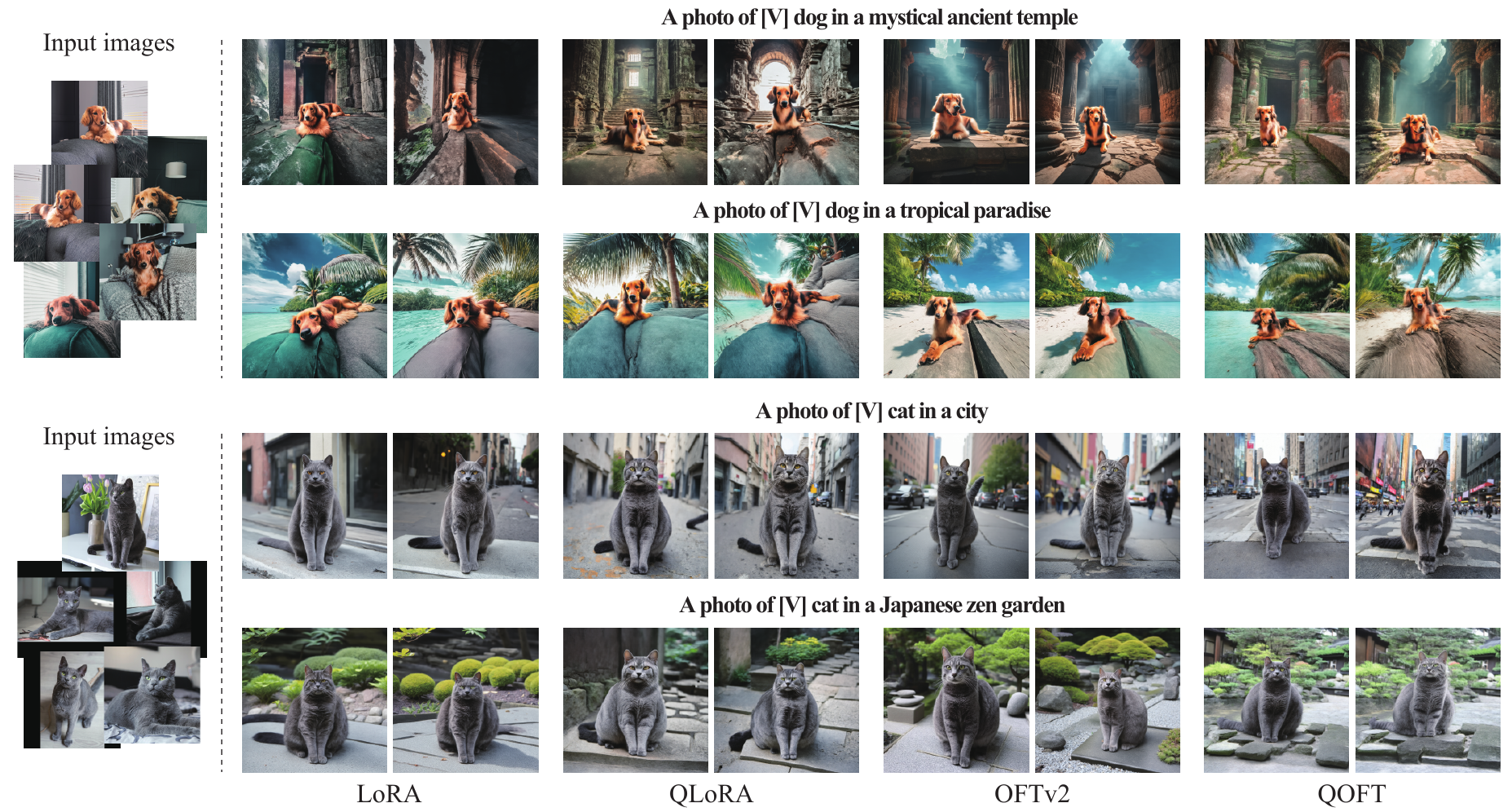}
    \caption{Qualitative results from Dreambooth fine-tuning of Stable Diffusion 3.5 Medium (8.1B parameters), with peak allocated GPU memory: LoRA (\textbf{38.00 GB}), OFT (\textbf{38.02 GB}), QLoRA (\textbf{35.03 GB}) and QOFT (\textbf{35.02 GB}).}\label{fig:sd3_medium}
\end{figure*}

Here we provide additional qualitative results of fine-tuning the Stable Diffusion 3.5 Medium model in Figure~\ref{fig:sd3_medium}.

The actual GPU memory usage during LoRA and OFTv2 fine-tuning is summarized in Table~\ref{tab:mem_sd}. As shown, OFTv2/QOFT demonstrates memory efficiency similar to LoRA and QLoRA, regardless of data precision or model scale.

\begin{table*}[h]
    \small
    \setlength{\tabcolsep}{9pt}
    \renewcommand{\arraystretch}{1.4}
    \centering
    \begin{tabular}{l|cc}
     & \textbf{SD 3.5 Medium} & \textbf{SD 3.5 Large} \\
    \shline
    LoRA  & 38.00 GB & 52.33 GB \\\rowcolor{Gray}
    OFTv2   & 38.02 GB & 52.32 GB   \\
    \hline
    QLoRA & 35.03 GB & 41.60 GB   \\\rowcolor{Gray}
    QOFT  & 35.02 GB & 41.53 GB   \\
    \end{tabular}
    \caption{Actual GPU memory usage during fine-tuning: LoRA, QLoRA, OFTv2, and QOFT applied on Stable Diffusion 3.5 Medium and Large.}
    \label{tab:mem_sd}
\end{table*}

\end{document}